\documentclass{article}
\usepackage{iclr2027_conference,times}

\usepackage{amsmath,amsfonts,bm}

\def\eqref#1{equation~\ref{#1}}

\def\1{\bm{1}}

\DeclareMathAlphabet{\mathsfit}{\encodingdefault}{\sfdefault}{m}{sl}
\SetMathAlphabet{\mathsfit}{bold}{\encodingdefault}{\sfdefault}{bx}{n}

\usepackage{hyperref}
\hypersetup{hidelinks}
\usepackage{url}
\usepackage{booktabs}
\usepackage{amsmath}
\usepackage{amssymb}
\usepackage{graphicx}
\usepackage{wrapfig}
\usepackage{enumitem}
\usepackage{capt-of}

\title{LOCKR: A Hidden-State Trajectory-Guided Planner for Detecting and Repairing Stable-but-Wrong Lock-In in Diffusion Language Models}

\author{
Guoshenghui Zhao \\
Rochester Institute of Technology \\
\texttt{gz1626@rit.edu}
\And
Tan Yu \\
NVIDIA Corporation \\
\texttt{tayu@nvidia.com}
\And
Weijie Zhao \\
Rochester Institute of Technology \\
\texttt{wjz@cs.rit.edu}
}

\iclrfinalcopy

\begin{document}

\maketitle
\lhead{}

\begin{abstract}
Diffusion language models generate text through iterative denoising, exposing intermediate trajectories before final answers are produced. We identify a recurring reasoning failure, \emph{stable-but-wrong lock-in}, where an answer stabilizes early around an incorrect value while substantial denoising remains. Surface-level decoding signals such as confidence, entropy, margin, and answer stability are insufficient to reliably distinguish correct from erroneous lock-in. We formulate selective reasoning repair as a lightweight test-time planning problem and propose \textsc{LOCKR}, a hidden-state trajectory-guided planner that decides when to allocate additional computation, expands a structured set of targeted repair branches, and selects the most promising continuation using trajectory-aware verification. Across two diffusion language models and three mathematical reasoning benchmarks, hidden-state trajectories consistently outperform surface signals and single hidden snapshots for both wrong-lock-in detection and repair selection. On natural evaluation distributions, \textsc{LOCKR} yields absolute accuracy gains of $2.21$--$5.37$ percentage points across all five evaluated settings, with repair rates ranging from $22\%$ to $41\%$. These results establish hidden diffusion trajectories as actionable signals for selective test-time reasoning repair.
\end{abstract}

\section{Introduction}

\begin{wrapfigure}{r}{0.50\columnwidth}
\vspace{-12pt}
\centering
\includegraphics[width=\linewidth]{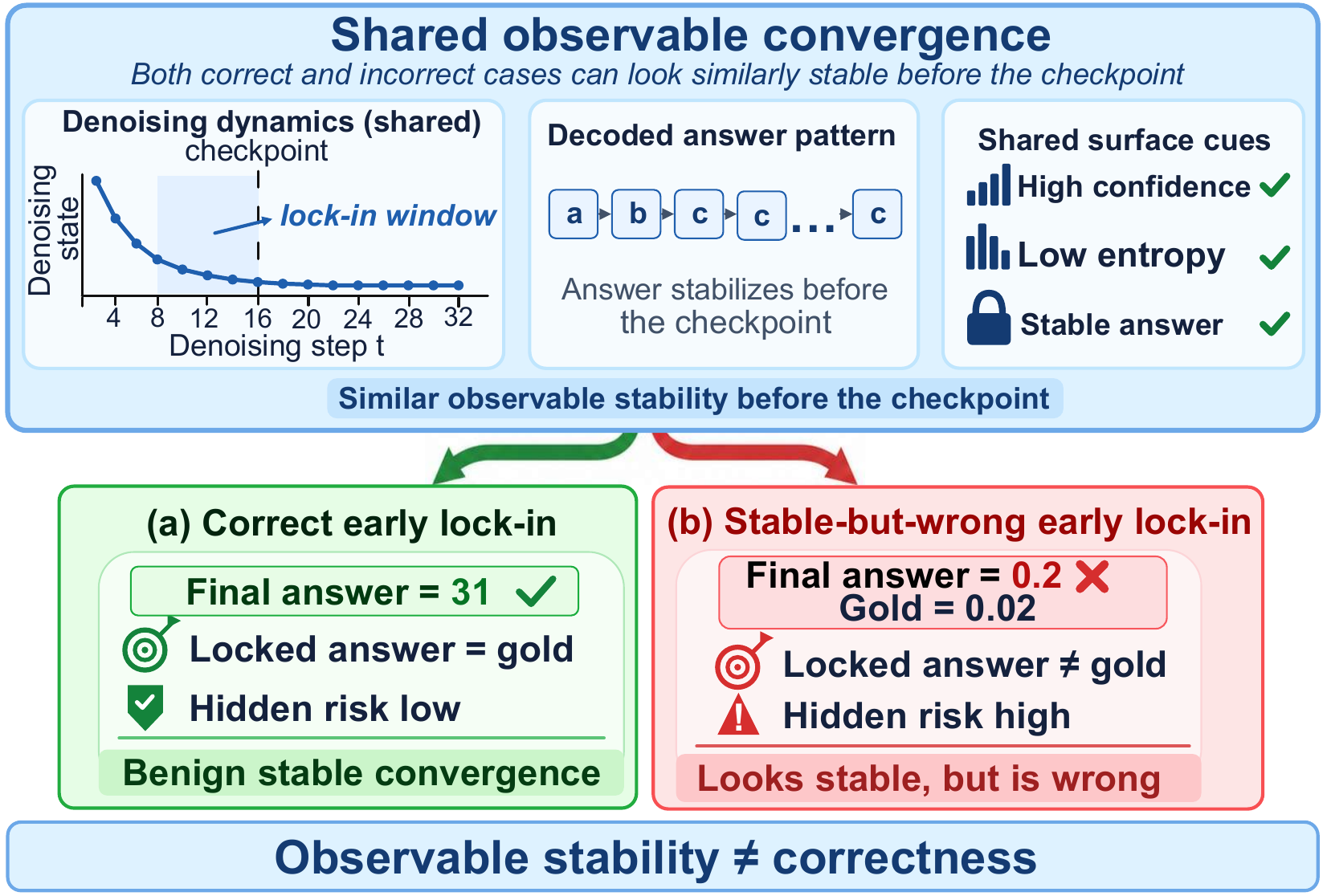}
\caption{\textbf{Stable-but-wrong early lock-in.} Correct and erroneous trajectories can exhibit similarly stable answers and reassuring surface signals before the midpoint checkpoint. The shaded region denotes the lock-in window; observable stability does not imply correctness.}
\label{fig:sbw_phenomenon}
\vspace{-18pt}
\end{wrapfigure}

Diffusion language models (dLLMs) do not simply produce an answer; they expose the process by which an answer emerges, stabilizes, and is repeatedly revised through iterative denoising \citep{li2022diffusionlm,han2023ssdlm,nie2025llada,ye2025dream}. This makes their intermediate trajectories unusually accessible compared with standard left-to-right generation, creating an opportunity to detect reasoning failures before the final output is reached. Yet this opportunity comes with a subtle challenge: an answer can appear to have converged while still being wrong.

We use \emph{lock-in} to describe a locally stable phase of denoising in which the same extractable answer persists across a sustained window of intermediate steps, even though generation has not yet finished. A lock-in can be benign when the stabilized answer is correct, or erroneous when it is incorrect. We refer to the latter case as \emph{stable-but-wrong} (SBW) lock-in. Figure~\ref{fig:sbw_phenomenon} illustrates the central ambiguity: a correct trajectory and an erroneous trajectory can exhibit similarly persistent answers and reassuring surface signals before the same intermediate checkpoint. In other words, observable convergence does not by itself reveal whether the model has converged to the right answer.

This ambiguity makes surface-level reliability signals potentially misleading. Observable decoding statistics such as confidence, entropy, logit margin, and answer stability are widely used as proxies for model reliability, but they can remain imperfect indicators of correctness \citep{guo2017calibration,jiang2021calibration,kadavath2022know,manakul2023selfcheckgpt,farquhar2024semanticentropy}. To test whether hidden model states contain information beyond these cues, we construct surface-matched correct and wrong lock-in pairs that are closely balanced across decoding statistics. Under this controlled setting, hidden-state representations remain strongly predictive of erroneous lock-in, and modeling their evolution over denoising consistently outperforms using a single hidden-state snapshot.

Motivated by this observation, we formulate selective reasoning repair as a lightweight test-time planning problem and propose \textsc{LOCKR}\footnote{\href{https://anonymous.4open.science/r/LOCKR-30F6/README.md}{Anonymous code repository.}}, a hidden-state trajectory-guided planner. Given the observed denoising prefix, \textsc{LOCKR} treats the hidden trajectory as a planning state, decides whether additional computation should be allocated, expands a structured set of counterfactual repair actions when intervention is warranted, and evaluates the resulting trajectories to determine which continuation should be retained. Concretely, a trajectory-based risk detector chooses between continuing the current decoding path and entering repair; targeted remasking instantiates a fixed local action bank; and a trajectory-aware verifier and final selector rank the resulting branches. Rather than restarting generation or allocating extra computation uniformly, \textsc{LOCKR} plans where additional computation is spent and which counterfactual continuation is ultimately kept.

We evaluate \textsc{LOCKR} on two diffusion language models and three mathematical reasoning benchmarks. Hidden-state trajectories consistently outperform surface statistics and validation-selected single hidden snapshots for both wrong-lock-in detection and repair-branch selection. On natural evaluation distributions, \textsc{LOCKR} yields absolute accuracy gains of $2.21$--$5.37$ percentage points across all five evaluated model--dataset settings, with repair rates ranging from $22\%$ to $41\%$. We further find that successful repair depends not only on detecting erroneous lock-in, but also on whether the available repair branches contain a correct alternative, revealing repair-space coverage as an important limit on achievable gains.

Our contributions are threefold:
\begin{itemize}[leftmargin=1.2em,itemsep=0pt,parsep=0pt,topsep=2pt,partopsep=0pt]
\item We identify and characterize SBW lock-in as a failure mode of diffusion-language-model reasoning and introduce a surface-matched evaluation protocol for separating erroneous from correct lock-in under controlled observable decoding statistics.
\item We formulate selective repair as a local test-time planning problem and propose \textsc{LOCKR}, which uses hidden trajectories as state representations to allocate repair computation, expand a fixed counterfactual action bank, and select among the resulting continuations with trajectory-aware value estimates.
\item We show across multiple dLLMs and reasoning benchmarks that hidden-state trajectories provide stronger signals than surface statistics or single hidden snapshots for both failure detection and repair selection, leading to consistent end-to-end accuracy improvements with selective inference-time intervention.
\end{itemize}

\section{Related Work}

\noindent\textbf{Diffusion Language Models and Inference Dynamics.} Diffusion language modeling has progressed from continuous and discrete diffusion formulations to masked, blockwise, and large-scale text generation, with recent systems demonstrating increasingly capable generation, reasoning, and efficient inference \citep{sohl2015deep,ho2020ddpm,austin2021structured,li2022diffusionlm,han2023ssdlm,sahoo2024mdlm,gong2025scaling,arriola2025block,nie2025llada,ye2025dream,bie2025llada20,diffusiongemma2026}.

Recent work further exploits intermediate denoising dynamics for token ordering, caching, early commitment, and convergence-aware decoding \citep{wu2025fastdllm,liu2025dllmcache,wang2026d2f,wang2026timefeature,li2026prophet,mohamed2026sched,li2026adadlm}. These methods primarily ask when decoding has stabilized or can terminate efficiently. Our focus is complementary: we ask whether an apparently stable answer is correct, and use hidden denoising trajectories to distinguish correct convergence from SBW lock-in.

\noindent\textbf{Test-Time Scaling and Self-Correction.} Test-time computation improves reasoning through diverse sampling, verification, structured search, and iterative refinement \citep{wei2022cot,cobbe2021verifiers,lightman2024verify,wang2023selfconsistency,madaan2023selfrefine,yao2023tot,besta2024got,snell2025testtime,deepseek2025r1}. Diffusion models offer an additional correction mechanism because intermediate drafts can be revisited: ReMDM enables remasking-based refinement \citep{wang2025remdm}, while related methods explore full-trajectory refinement, self-reflective remasking, learned token-quality estimates, and trajectory search \citep{dang2025particlegibbs,huang2026remedi,kim2026prism,bai2026prism}. \textsc{LOCKR} differs by treating correction as selective local planning: prefix hidden trajectories determine whether to expand a repair space, and repair trajectories provide value signals for selecting among the resulting continuations.

\noindent\textbf{Internal Representations for Error Detection and Verification.} Calibration, self-evaluation, semantic uncertainty, and internal representations have all been studied as signals of output reliability \citep{guo2017calibration,jiang2021calibration,kadavath2022know,manakul2023selfcheckgpt,farquhar2024semanticentropy}. Internal-state methods further show that hidden geometry and cross-layer dynamics can reveal errors that output-level uncertainty misses \citep{chen2024inside,su2024mind,chen2024sharpness,zhang2025icr}. Most closely, HIVE \citep{zhao2026hive} aggregates hidden evidence across dLLM denoising steps for post-generation verification. \textsc{LOCKR} instead studies early lock-in under surface-matched conditions and turns hidden trajectories into an actionable planning signal that controls both intervention and repair-branch selection.

\section{SBW Lock-In in Diffusion Language Models}
\label{sec:sbw}

\subsection{Iterative Denoising and Early Lock-In}
\label{sec:early_lockin}

Consider a diffusion language model observed over $T$ recorded denoising steps. We denote the model state at step $t$ by $\mathbf{x}^{(t)}$, where $t\in\{0,\ldots,T-1\}$ and larger $t$ corresponds to a later stage of denoising. The observed trajectory is $\mathbf{x}^{(0)},\mathbf{x}^{(1)},\ldots,\mathbf{x}^{(T-1)}$. At each step, a deterministic answer-extraction function $\mathcal{E}$ yields the current normalized answer $a_t=\mathcal{E}(\mathbf{x}^{(t)})$, where $a_t=\emptyset$ means that no valid answer can yet be extracted.

We evaluate early lock-in at the midpoint checkpoint $c=\operatorname{round}(0.5(T-1))$ using a lock-in window of length $w=\max(1,\lceil(c+1)/2\rceil)$, corresponding to the most recent half of the answer states observed up to $c$. An example is \emph{early locked} if all answers in this window are valid and identical:
\[
a_{c-w+1}=a_{c-w+2}=\cdots=a_c=\ell\neq\emptyset,
\]
where $\ell$ denotes the locked answer.

Finally, let $g$ denote the gold answer and let $\mathcal{J}(\ell,g)\in\{0,1\}$ denote the task-specific correctness function. We call an early-locked example an \emph{early-correct lock-in} if $\mathcal{J}(\ell,g)=1$, and an \emph{early-wrong lock-in} if $\mathcal{J}(\ell,g)=0$. The label is determined at the intermediate checkpoint and does not require the answer to remain unchanged throughout the remaining denoising steps. This definition captures the setting of interest: the model already exhibits sustained answer agreement while substantial generation remains, yet the answer around which it has stabilized may be correct or incorrect.

\subsection{The SBW Phenomenon}
\label{sec:sbw_phenomenon}

We refer to early-wrong lock-in as SBW lock-in. The phenomenon highlights a fundamental ambiguity in iterative generation: observable convergence is not equivalent to correctness. Two generations may exhibit equally persistent answers before the same intermediate checkpoint, yet one has locked onto a correct answer while the other has locked onto an incorrect one.

SBW is common rather than exceptional, accounting for $14.9\%$--$60.5\%$ of processed generations across the five formal model--dataset settings. Appendix Table~\ref{tab:app_sbw_prevalence} reports the corresponding counts and setting-specific rates.

As illustrated earlier in Figure~\ref{fig:sbw_phenomenon}, correct and erroneous lock-ins can exhibit similarly stable answers and reassuring surface cues. Directly comparing arbitrary correct and wrong generations would therefore leave a confound: a detector might exploit differences in confidence, uncertainty, or surface stability rather than information specific to erroneous reasoning dynamics. This motivates controlling observable decoding behavior before testing whether hidden trajectories provide additional correctness information.

\subsection{Surface-Matched Benchmark Construction}
\label{sec:surface_matching}

\begin{wrapfigure}{r}{0.50\columnwidth}
\vspace{-14pt}
\centering
\includegraphics[width=\linewidth]{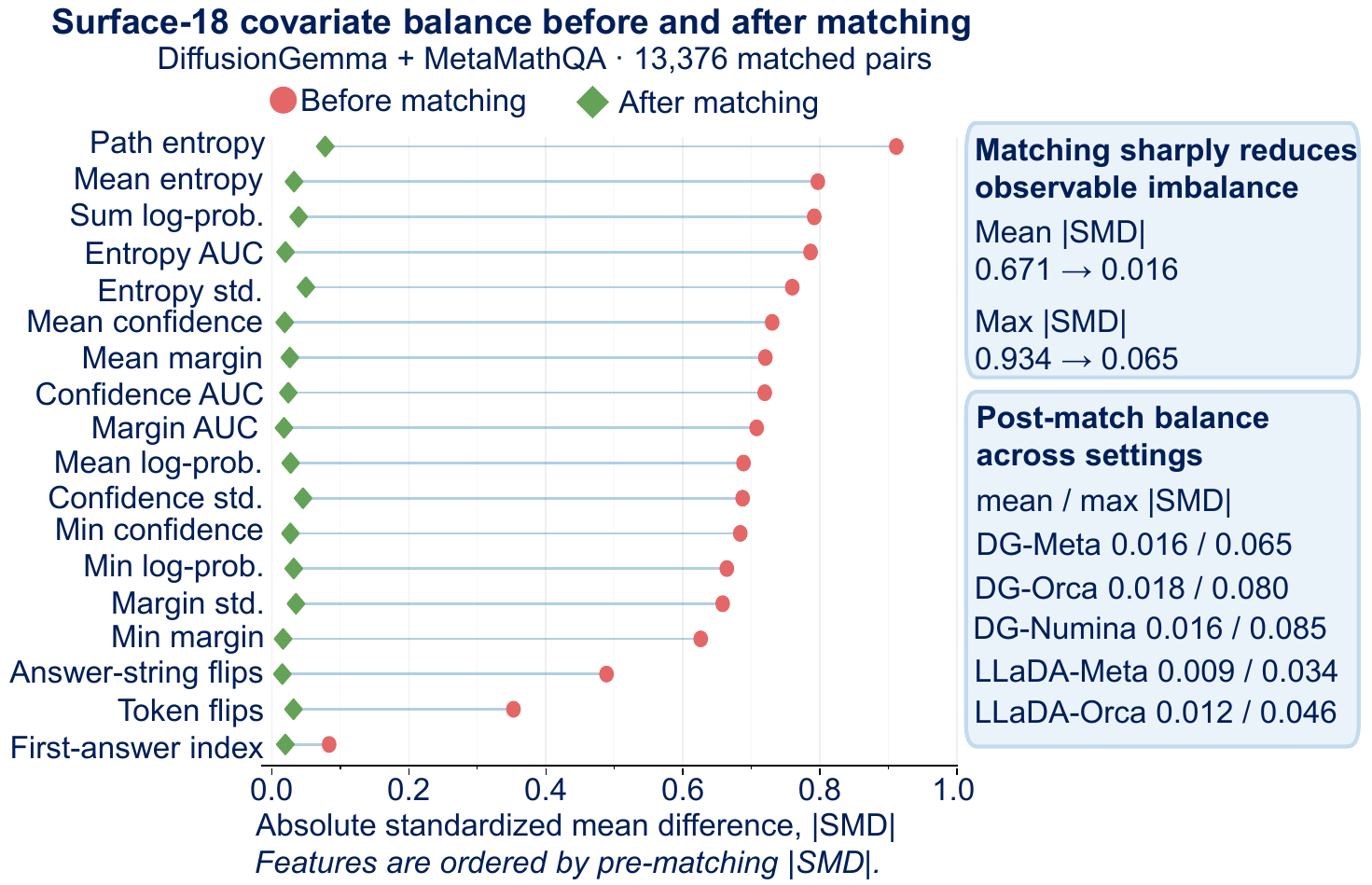}
\caption{\textbf{Surface-18 balance before and after matching.} On DiffusionGemma + MetaMathQA, matching reduces mean/max $|\mathrm{SMD}|$ from $0.671/0.934$ to $0.016/0.065$. The inset summarizes post-matching balance across all five formal settings, with every maximum below $0.085$.}
\label{fig:surface_balance}
\vspace{-18pt}
\end{wrapfigure}

To isolate erroneous lock-in from observable decoding differences, we construct a balanced benchmark of matched early-correct and early-wrong examples. Each generation is represented by a Surface-18 vector $\mathbf{s}\in\mathbb{R}^{18}$ comprising 18 observable prefix-level statistics that capture confidence, entropy, probability margin, token and answer stability, answer-emergence timing, log-probability behavior, and temporal summaries of these signals. Figure~\ref{fig:surface_balance} visualizes balance across these 18 dimensions, while Appendix~\ref{app:surface18_def} and Table~\ref{tab:surface18_definitions} provide the complete feature definitions.

Following matched-design principles \citep{rosenbaum1983propensity,stuart2010matching,austin2009balance}, we one-to-one match early-correct and early-wrong examples using Surface-18 after robust normalization, subject to feature-specific calipers and a weighted-distance constraint. Hidden representations are never used during matching; full construction details are provided in Appendix~\ref{app:matching_algorithm}. Post-matching balance is assessed with the mean and maximum absolute standardized mean difference (SMD) across the 18 features (Appendix~\ref{app:balance_stats}).

\begin{wraptable}{r}{0.45\columnwidth}
\vspace{-20pt}
\centering
\scriptsize
\setlength{\tabcolsep}{2pt}
\caption{\textbf{Surface-matched benchmark.} Src. Acc. denotes decoding accuracy in the source population before matching; lower $|\mathrm{SMD}|$ indicates stronger balance.}
\label{tab:sbw_matching}
\begin{tabular*}{\linewidth}{@{\extracolsep{\fill}}lrrc@{}}
\toprule
\textbf{Setting} & \textbf{Src. Acc. (\%)} & \textbf{Pairs} & \textbf{Mean / Max $|\mathrm{SMD}|$} \\
\midrule
DG-Meta & 79.21 & 13,376 & .01629 / .06530 \\
DG-Orca & 70.28 & 16,067 & .01765 / .08035 \\
DG-Numina & 37.60 & 19,137 & .01622 / .08494 \\
LLaDA-Meta & 42.93 & 25,064 & .00947 / .03416 \\
LLaDA-Orca & 40.49 & 22,418 & .01186 / .04639 \\
\bottomrule
\end{tabular*}
\vspace{-12pt}
\end{wraptable}

Figure~\ref{fig:surface_balance} shows that, for DiffusionGemma + MetaMathQA, mean/max absolute SMD falls from $0.671/0.934$ before matching to $0.016/0.065$ afterward. Table~\ref{tab:sbw_matching} reports the exact benchmark sizes and confirms a maximum absolute SMD below $0.085$ in every formal setting.

The matched benchmark therefore suppresses much of the marginal surface separation between correct and erroneous lock-in. Because low SMD does not imply identical multivariate distributions, learned Surface-18 classifiers remain meaningful controls. This tests whether hidden trajectories distinguish correct from erroneous lock-in after observable behavior is matched.

\section{LOCKR: Hidden-State Trajectory-Guided Test-Time Planning}
\label{sec:lockr}

\begin{figure*}[t!]
\vspace{-12pt}
\centering
\includegraphics[width=0.95\textwidth]{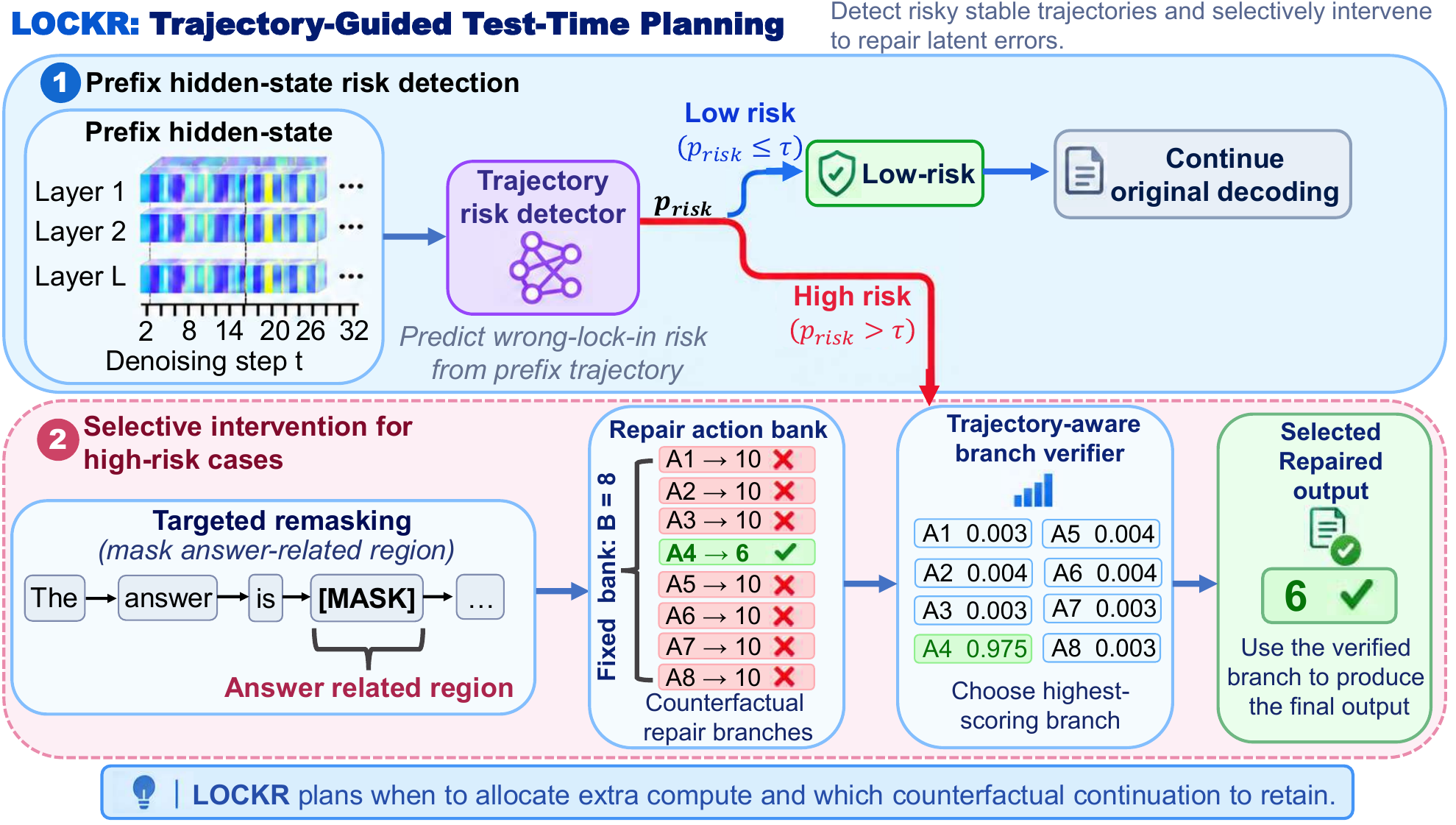}
\caption{\textbf{Trajectory-guided test-time planning with \textsc{LOCKR}.} The observed prefix hidden trajectory serves as the planning state. Low-risk generations retain the baseline path, while high-risk generations trigger targeted remasking with a fixed repair-action bank. The resulting counterfactual trajectories are scored by a trajectory-aware verifier and final selector, which retains the highest-valued continuation.}
\label{fig:lockr_overview}
\vspace{-14pt}
\end{figure*}

\subsection{Overview}
\label{sec:lockr_overview}

\textsc{LOCKR} treats selective reasoning repair as a lightweight test-time planning problem (Figure~\ref{fig:lockr_overview}). The observed prefix hidden trajectory forms the planning state: a risk model either retains the baseline path or triggers targeted remasking to expand a local counterfactual repair space. The resulting repair trajectories are then scored to select the most promising continuation.

A repair action specifies both \emph{where} to intervene along the stored baseline trajectory and \emph{how strongly} to reopen the current draft. We define $\mathcal{A}_i=(\rho_i,K_i)$, where $\rho_i$ selects an intermediate baseline state and $K_i$ is the nominal remasking budget applied to an answer-related region before denoising resumes. Executing $\mathcal{A}_i$ produces one counterfactual continuation. A fixed bank of $B$ actions spans multiple intervention locations and remasking strengths and is specified before training rather than learned or adapted per example. The experimental instantiation is given in Section~\ref{sec:exp_setup}, with exact action pairs and model-specific remasking procedures in Appendices~\ref{app:repair_action_bank} and~\ref{app:targeted_remasking}.

Formally, let $b_0$ denote the continuation produced by retaining the original decoding path, let $\mathcal{A}=\{\mathcal{A}_1,\ldots,\mathcal{A}_B\}$ denote the fixed local repair-action bank, and let $\mathcal{B}(\mathbf{x};\mathcal{A})$ denote the repair branches obtained by executing these actions from the available baseline states. Conditional on an example satisfying the early-lock-in eligibility criterion, the planning policy is
\[
\pi_{\textsc{LOCKR}}(\mathbf{H},\mathbf{x})=
\begin{cases}
b_0, & p_{\mathrm{risk}}\leq\tau,\\[3pt]
\displaystyle \arg\max_{b_i\in\mathcal{B}(\mathbf{x};\mathcal{A})} s_i, & p_{\mathrm{risk}}>\tau,
\end{cases}
\]
where $s_i$ is the final correctness value assigned to branch $b_i$. Thus, prefix trajectories determine whether repair is invoked, the fixed action bank defines the candidate futures, and the verifier/selector determines which continuation is retained. The planner learns \emph{when} additional computation is warranted and \emph{which} resulting continuation to retain; it does not learn the repair action bank itself.

\subsection{Hidden-State Trajectory Encoding and Risk Detection}
\label{sec:lockr_risk}

The observable answer trajectory used to define lock-in captures only the model's decoded behavior. To access information that may remain hidden despite similar surface behavior, \textsc{LOCKR} records internal representations throughout the prefix leading to the lock-in checkpoint. Let $\mathbf{h}_{t,l,p}$ denote the hidden representation at denoising step $t$, model layer $l$, and token position $p$. We collect representations from multiple relative denoising positions and multiple model layers, retaining both a fixed answer-focused token track and a complementary change track that records positions whose tokens change between successive selected snapshots.

Because raw hidden activations are high dimensional, we compress selected states using fixed layer-specific random-sign projections \citep{achlioptas2003random} and order them across denoising positions, layers, and answer/change token tracks to form the hidden trajectory $\mathbf{H}$. This representation preserves temporal and localized internal changes; the full extraction geometry is provided in Appendix~\ref{app:hidden_trajectory}.

A bidirectional GRU \citep{cho2014gru,schuster1997bilstm} encodes $\mathbf{H}$ and predicts early-lock-in correctness. We define wrong-lock-in risk as $p_{\mathrm{risk}}=1-f_{\theta}(\mathbf{H})$ and use a validation-selected threshold $\tau$ to retain the baseline path or trigger repair. The risk model uses only hidden states observed no later than the early-lock-in checkpoint, so later denoising states cannot revise the intervention decision.

\subsection{Targeted Remasking and Repair Branch Generation}
\label{sec:lockr_repair}

The lower stage of Figure~\ref{fig:lockr_overview} converts a high-risk generation into a structured repair pool. Rather than restarting from noise, \textsc{LOCKR} reuses intermediate drafts from the baseline trajectory, locally remasks an answer-related region, and resumes denoising, thereby preserving most of the existing draft while reopening part of the sequence for revision.

For each high-risk generation, a fixed bank of $B$ paired checkpoint--remasking actions expands the current state into counterfactual repair branches $\mathcal{B}=\{b_1,\ldots,b_B\}$. The actions vary intervention time and strength rather than repeatedly sampling from one identical state; exact checkpoint fractions, remasking budgets, and model-specific procedures are provided in Appendix~\ref{app:repair_action_bank} and Appendix~\ref{app:targeted_remasking}. Repair may start from baseline states before or after the early-lock-in checkpoint, but the intervention decision remains fixed by the prefix-level risk estimate in Section~\ref{sec:lockr_risk}; later states are used only as repair starting points.

\subsection{Trajectory-Aware Branch Verification and Selection}
\label{sec:lockr_verification}

For each repair branch $b_i$, we extract a hidden trajectory $\mathbf{H}^{\mathrm{br}}_i$ using the same multi-step, multi-layer construction as the first-stage detector, but over the repair continuation. A separate trajectory verifier maps this representation to a branch score $v_i=f_{\phi}^{\mathrm{br}}(\mathbf{H}^{\mathrm{br}}_i)$; higher scores indicate stronger evidence that the candidate is correct. The verifier is trained with auxiliary correctness, pairwise-ranking, and listwise objectives, with full architecture and training details deferred to Appendix~\ref{app:branch_verifier}.

The frozen trajectory score is then combined with the original risk context, within-pool answer agreement, and lightweight repair metadata using a histogram gradient-boosted selector \citep{friedman2001gbm}. This selector produces a final score $s_i$ and returns $b^{\star}=\arg\max_{b_i\in\mathcal{B}}s_i$. Prefix trajectories therefore determine whether repair is invoked, while repair trajectories help determine which counterfactual continuation is retained. Full selector features and hyperparameters are provided in Appendix~\ref{app:hgb_selector}.

\section{Experiments}
\label{sec:experiments}

\subsection{Experimental Setup}
\label{sec:exp_setup}

\noindent\textbf{Models and datasets.} We evaluate \textsc{LOCKR} on DiffusionGemma \citep{diffusiongemma2026} and LLaDA-2 \citep{bie2025llada20} using MetaMathQA \citep{yu2024metamath}, Orca-Math \citep{mitra2024orcamath}, and NuminaMath V2 \citep{numinamath2024}. DiffusionGemma is evaluated on all three benchmarks and LLaDA-2 on MetaMathQA and Orca-Math. Unless otherwise specified, base generation uses 32 denoising steps, a maximum generation length of 256 tokens, and the midpoint early-lock-in checkpoint from Section~\ref{sec:early_lockin}.

\noindent\textbf{Evaluation protocols.} We use two protocols: the balanced surface-matched benchmark from Section~\ref{sec:surface_matching} for controlled detection, branch-selection, and end-to-end analysis, and each model--dataset's original distribution for measuring overall accuracy under selective intervention. Because each matched pair contains one early-correct and one early-wrong lock-in, the controlled benchmark has a $50\%$ checkpoint-level baseline accuracy by construction. For end-to-end evaluation on this controlled benchmark, the baseline label is the frozen checkpoint-level correctness label used to construct each matched pair; repaired branches are evaluated with the same task-specific correctness function.

\noindent\textbf{Training and model selection.} All learned components, snapshot choices, and intervention operating points are selected using training/validation data only and frozen before held-out test evaluation. The risk detector and branch verifier are trained separately; full split, optimization, and model-selection details are provided in Appendix~\ref{app:training_validation}.

\noindent\textbf{Repair action bank.} In all experiments, we instantiate the repair bank with $B=8$ fixed checkpoint--remasking actions. We use eight actions as a fixed breadth--compute design point that covers multiple intervention times and remasking strengths while keeping branch expansion bounded. The bank is fixed before training, shared unchanged across all model--dataset settings, and is neither learned nor selected using validation or test performance. The exact action pairs are listed in Appendix~\ref{app:repair_action_bank}.

\noindent\textbf{Baselines.} We compare against surface-only, confidence/self-consistency, remasking, external PRM \citep{yang2024qwen25math,zheng2024processbench,zhang2025prmlessons}, and validation-selected single-snapshot baselines, using identical repair pools where applicable. Oracle quantities are analysis-only upper bounds; complete baseline definitions are provided in Appendix~\ref{app:training_validation}.

\noindent\textbf{Metrics.} We report AUROC/pairwise accuracy for detection, selected-branch accuracy for repair selection, and final accuracy with relative denoising cost for end-to-end evaluation; natural-distribution results additionally report repair rate. Standard decoding has cost $1.0\times$; full cost accounting is given in Appendix~\ref{app:cost_accounting}.

\subsection{Detecting SBW Lock-In}
\label{sec:exp_detection}

We first evaluate whether the observed prefix can identify SBW lock-in before generation completes. All results use the surface-matched benchmark from Section~\ref{sec:surface_matching}, which controls observable decoding statistics before hidden-state comparison. This design tests whether hidden trajectories provide correctness information beyond matched confidence, uncertainty, and stability cues.

\begin{wrapfigure}[9]{r}{0.44\columnwidth}
\vspace{-16pt}
\centering
\includegraphics[width=\linewidth]{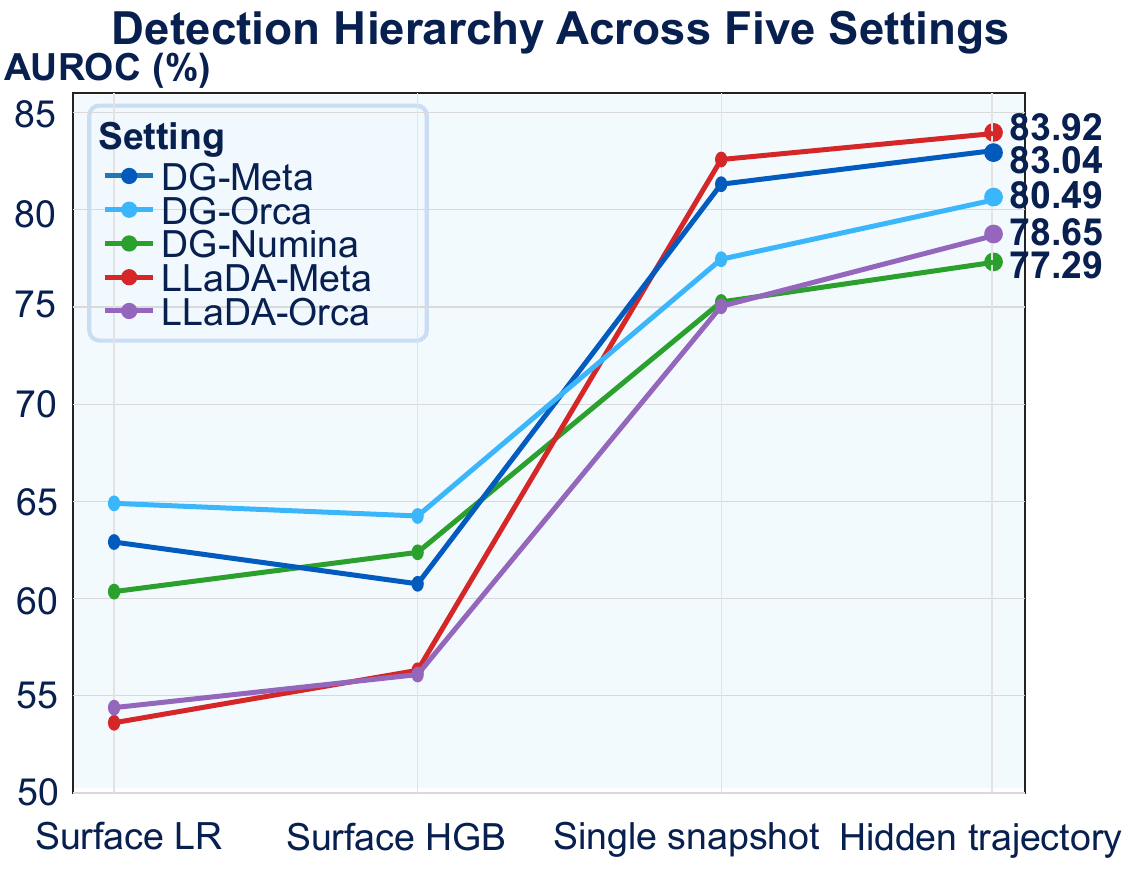}
\caption{\textbf{Detection hierarchy across five settings.} Hidden representations outperform Surface-18 baselines, and full trajectories improve over the validation-selected snapshot in all five settings.}
\label{fig:detection_hierarchy}
\vspace{-10pt}
\end{wrapfigure}

\noindent
\begin{minipage}[t]{0.52\columnwidth}
\vspace{-18pt}
\centering
\scriptsize
\setlength{\tabcolsep}{2pt}
\captionof{table}{\textbf{Wrong-lock-in AUROC (\%).} S-LR/S-HGB denote Surface-18 logistic regression/histogram gradient boosting; Snap. is the validation-selected hidden snapshot.}
\label{tab:wrong_lockin_detection}
\begin{tabular*}{\linewidth}{@{\extracolsep{\fill}}lrrrr@{}}
\toprule
\textbf{Setting} & \textbf{S-LR} & \textbf{S-HGB} & \textbf{Snap.} & \textbf{\textsc{LOCKR}} \\
\midrule
DG-Meta & 62.91 & 60.76 & 81.32 & \textbf{83.04} \\
DG-Orca & 64.90 & 64.25 & 77.46 & \textbf{80.49} \\
DG-Numina & 60.36 & 62.38 & 75.26 & \textbf{77.29} \\
LLaDA-Meta & 53.61 & 56.31 & 82.59 & \textbf{83.92} \\
LLaDA-Orca & 54.39 & 56.09 & 75.04 & \textbf{78.65} \\
\bottomrule
\end{tabular*}
\end{minipage}

\par\smallskip

Table~\ref{tab:wrong_lockin_detection} summarizes the principal AUROC comparison, while Figure~\ref{fig:detection_hierarchy} visualizes the hierarchy across representation families. Individual scalar surface signals remain close to chance across all settings; the complete AUROC and pairwise-accuracy results, including these scalar baselines, are reported in Appendix Table~\ref{tab:app_wrong_lockin_detection_full}. Combining the 18 observable statistics provides a stronger signal, with Surface-18 LR and HGB reaching $53.61$--$64.90\%$ AUROC, but a substantial gap remains relative to hidden representations. A validation-selected single hidden snapshot already provides strong discrimination, reaching $75.04$--$82.59\%$ AUROC.

The final transition in Figure~\ref{fig:detection_hierarchy} isolates the value of temporal hidden-state evolution: all five model--dataset curves rise from the single-snapshot baseline to the complete hidden trajectory. \textsc{LOCKR} reaches $77.29$--$83.92\%$ AUROC and $78.24$--$84.08\%$ pairwise accuracy. Relative to the validation-selected single hidden snapshot, trajectory modeling improves AUROC by $1.33$--$3.61$ percentage points and pairwise accuracy by $1.41$--$4.75$ percentage points. Thus, access to hidden representations accounts for the largest improvement over surface-only detection, while modeling their evolution supplies a smaller but consistent additional signal beyond a static internal state.

\subsection{Selecting Successful Repair Branches}
\label{sec:exp_branch_selection}

We next isolate branch valuation: given the same fixed repair pool, can hidden trajectories distinguish successful counterfactual futures from unsuccessful ones? All methods rank identical candidate sets, so differences reflect selection quality rather than candidate generation. Appendix Figure~\ref{fig:qual_branch_selection} illustrates a representative case in which the unique correct branch is outvoted $7$-to-$1$ but recovered by trajectory-aware selection.

\begin{wraptable}{r}{0.45\columnwidth}
\vspace{-8pt}
\centering
\scriptsize
\setlength{\tabcolsep}{2pt}
\caption{\textbf{Fixed-pool branch-selection accuracy (\%).} All methods rank the same repair candidates.}
\label{tab:branch_selection}
\begin{tabular*}{\linewidth}{@{\extracolsep{\fill}}lrrrr@{}}
\toprule
\textbf{Setting} & \textbf{S-18} & \textbf{PRM} & \textbf{Snap.} & \textbf{\textsc{LOCKR}} \\
\midrule
DG-Meta & 59.6 & 62.3 & 64.5 & \textbf{67.8} \\
DG-Orca & 60.2 & 60.1 & 62.9 & \textbf{66.2} \\
DG-Numina & 52.4 & 52.5 & 53.6 & \textbf{55.2} \\
LLaDA-Meta & 44.0 & 48.8 & 51.7 & \textbf{56.0} \\
LLaDA-Orca & 42.3 & 45.9 & 47.7 & \textbf{50.8} \\
\bottomrule
\end{tabular*}
\vspace{-14pt}
\end{wraptable}

Table~\ref{tab:branch_selection} summarizes the principal fixed-pool selector comparison, while Appendix Table~\ref{tab:app_branch_selection_full} reports the complete results, including random selection, maximum confidence, and the oracle branch-pool upper bound. Simple output-level criteria provide limited guidance: maximum confidence performs similarly to random selection, while Surface-18 yields moderate improvements. The external PRM is stronger, but hidden representations are substantially more effective: the single hidden snapshot verifier improves over the surface and PRM baselines across all five settings.

The trajectory-aware \textsc{LOCKR} selector performs best among all non-oracle methods in every setting, reaching $55.2$--$67.8\%$ selected-branch accuracy and improving over the single hidden snapshot verifier by $1.6$--$4.3$ percentage points. These results show that generating alternatives and valuing them are distinct problems: repair trajectories provide useful information for ranking futures that surface agreement alone cannot resolve.

\subsection{End-to-End Test-Time Repair}
\label{sec:exp_e2e}

\begin{wrapfigure}[4]{r}{0.65\columnwidth}
\vspace{-14pt}
\centering
\includegraphics[width=\linewidth]{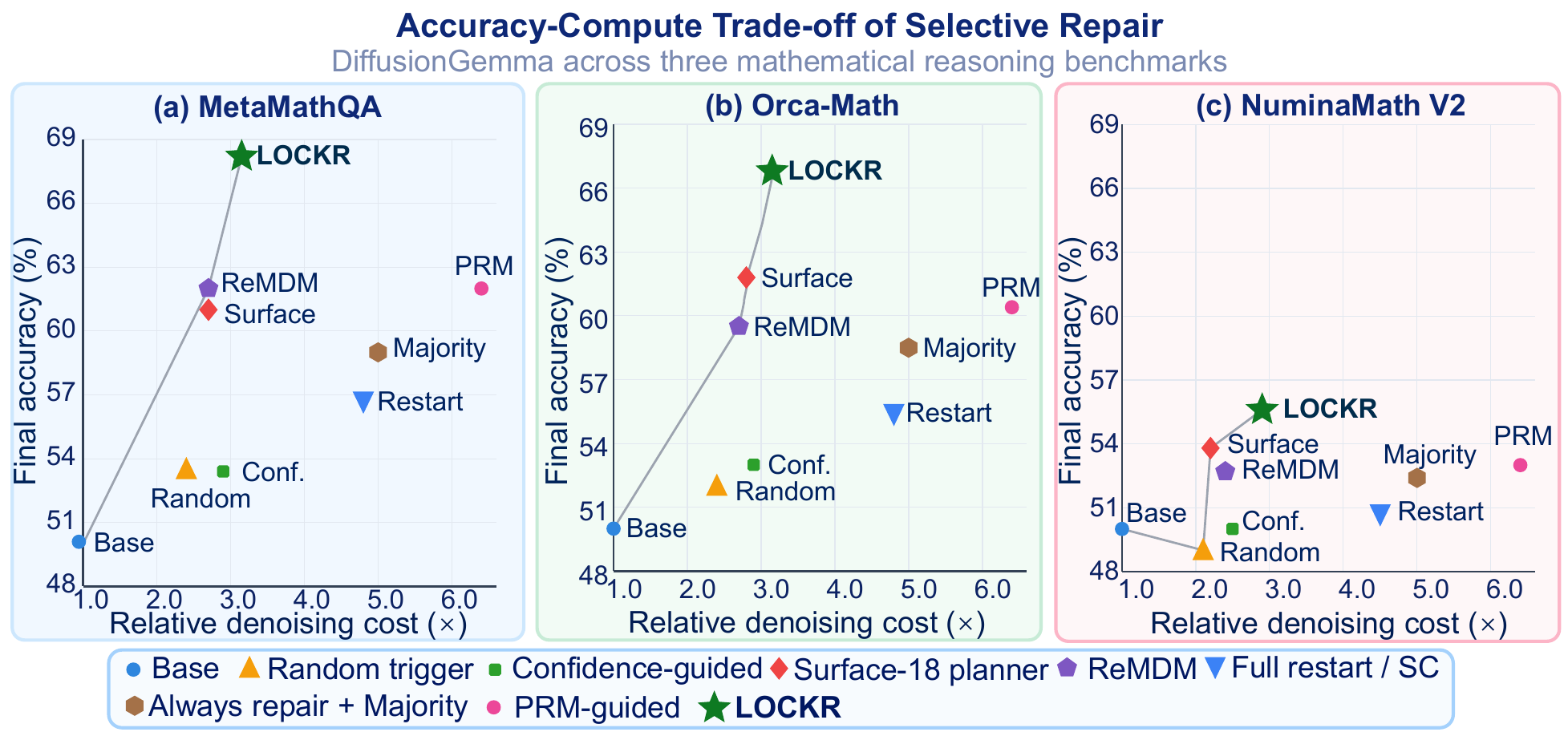}
\caption{\textbf{Accuracy--compute trade-off.} \textsc{LOCKR} achieves the highest non-oracle accuracy across all three DiffusionGemma settings at $2.9$--$3.1\times$ relative denoising cost. Full results are reported in Appendix Table~\ref{tab:app_e2e_diffusiongemma_full}.}
\label{fig:accuracy_compute_tradeoff}
\vspace{-10pt}
\end{wrapfigure}

We next evaluate the complete policy on the balanced surface-matched benchmark. Table~\ref{tab:e2e_diffusiongemma} and Figure~\ref{fig:accuracy_compute_tradeoff} summarize the DiffusionGemma accuracy--compute results, with full comparisons in Appendix Table~\ref{tab:app_e2e_diffusiongemma_full}.

\noindent
\begin{minipage}[t]{0.33\columnwidth}
\vspace{-16pt}
\centering
\scriptsize
\setlength{\tabcolsep}{0.8pt}
\captionof{table}{\textbf{End-to-end repair on DiffusionGemma.} Accuracy (\%) / relative denoising cost. Complete results are in Appendix Table~\ref{tab:app_e2e_diffusiongemma_full}.}
\label{tab:e2e_diffusiongemma}
\begin{tabular*}{\linewidth}{@{\extracolsep{\fill}}lccc@{}}
\toprule
\textbf{Method} & \textbf{Meta} & \textbf{Orca} & \textbf{Numina} \\
\midrule
Base & 50.0/1.0 & 50.0/1.0 & 50.0/1.0 \\
Random & 53.2/2.4 & 51.6/2.4 & 48.7/2.2 \\
Self-Cons. & 56.5/4.8 & 55.2/4.8 & 50.4/4.5 \\
Surface-18 & 60.8/2.7 & 61.1/2.8 & 53.6/2.2 \\
ReMDM & 61.3/2.7 & 59.2/2.7 & 52.4/2.4 \\
PRM & 61.9/6.4 & 60.2/6.4 & 52.8/6.2 \\
\textbf{\textsc{LOCKR}} & \textbf{67.9/3.1} & \textbf{66.4/3.1} & \textbf{55.4/2.9} \\
E2E oracle & 72.8/-- & 72.9/-- & 62.3/-- \\
\bottomrule
\end{tabular*}
\end{minipage}

\par\smallskip

Naive additional computation remains well below \textsc{LOCKR}. The complete planner reaches $67.9\%$, $66.4\%$, and $55.4\%$ accuracy, exceeding the strongest non-\textsc{LOCKR} baseline by $6.0$, $5.3$, and $1.8$ percentage points on MetaMathQA, Orca-Math, and NuminaMath V2, respectively.

\textsc{LOCKR} is Pareto-efficient in all three settings at $2.9$--$3.1\times$ relative cost. It improves both accuracy and cost over PRM-guided repair and always-repair majority voting on MetaMathQA and Orca-Math; on NuminaMath V2, it raises the attainable accuracy--compute envelope beyond the cheaper Surface-18 operating point. The gain therefore reflects selective allocation and branch valuation rather than uniformly increased test-time computation.

\subsection{Results Across Models and Datasets}
\label{sec:exp_generalization}

\begin{wraptable}{r}{0.40\columnwidth}
\vspace{-44pt}
\centering
\scriptsize
\setlength{\tabcolsep}{3pt}
\caption{\textbf{Evaluation on LLaDA-2.} Each entry reports accuracy (\%) / relative denoising cost.}
\label{tab:generalization_llada}
\begin{tabular*}{\linewidth}{@{\extracolsep{\fill}}lcc@{}}
\toprule
\textbf{Method} & \textbf{MetaMathQA} & \textbf{Orca-Math} \\
\midrule
Base decoding & 50.0 / 1.0 & 50.0 / 1.0 \\
Surface-18 planner & 44.9 / 2.3 & 42.6 / 2.0 \\
ReMDM & 48.5 / 2.4 & 45.8 / 2.3 \\
PRM-guided repair & 51.2 / 6.1 & 50.8 / 5.9 \\
Single hidden snapshot & 53.7 / 2.6 & 50.1 / 2.4 \\
\textbf{\textsc{LOCKR}} & \textbf{57.2 / 2.9} & \textbf{51.9 / 2.5} \\
E2E oracle & 71.7 / -- & 65.9 / -- \\
\bottomrule
\end{tabular*}
\vspace{-14pt}
\end{wraptable}

To test whether the gains persist across model families, we repeat the end-to-end evaluation with LLaDA-2 (Table~\ref{tab:generalization_llada}). This setting is harder for repair: several surface- or remasking-based baselines fail to improve over the $50\%$ matched-set baseline. \textsc{LOCKR} nevertheless gives the best non-oracle accuracy on both datasets and preserves the same ordering observed with DiffusionGemma: trajectory modeling outperforms surface-only planning and the validation-selected single snapshot. It reaches $57.2\%$ accuracy at $2.9\times$ cost on MetaMathQA and $51.9\%$ at $2.5\times$ cost on Orca-Math. The remaining gain varies with model, task, and repair-space headroom.

\subsection{Natural-Distribution Evaluation}
\label{sec:exp_natural}

\begin{wraptable}{r}{0.45\columnwidth}
\vspace{-22pt}
\centering
\scriptsize
\setlength{\tabcolsep}{2pt}
\caption{\textbf{Natural-distribution evaluation.} Repair rate is measured over all examples; cost is relative denoising work. ``pp'' denotes percentage points.}
\label{tab:natural_distribution}
\begin{tabular*}{\linewidth}{@{\extracolsep{\fill}}lccccc@{}}
\toprule
\textbf{Setting} & \textbf{Base (\%)} & \textbf{\textsc{LOCKR} (\%)} & \textbf{$\Delta$ (pp)} & \textbf{Rate} & \textbf{Cost} \\
\midrule
DG-Meta & 79.21 & \textbf{82.90} & \textbf{+3.69} & 22\% & 1.8 \\
DG-Orca & 70.28 & \textbf{74.50} & \textbf{+4.22} & 32\% & 2.2 \\
DG-Numina & 37.60 & \textbf{41.80} & \textbf{+4.20} & 40\% & 2.6 \\
LLaDA-Meta & 42.93 & \textbf{48.30} & \textbf{+5.37} & 35\% & 2.4 \\
LLaDA-Orca & 40.49 & \textbf{42.70} & \textbf{+2.21} & 41\% & 2.6 \\
\bottomrule
\end{tabular*}
\vspace{-14pt}
\end{wraptable}

We next apply the frozen validation-selected policy to each original model--dataset distribution, repairing only early-locked examples above the risk threshold. Table~\ref{tab:natural_distribution} shows consistent gains outside the controlled benchmark: \textsc{LOCKR} yields accuracy gains of $2.21$--$5.37$ percentage points across all five settings while repairing only $22$--$41\%$ of examples, corresponding to $1.8$--$2.6\times$ relative cost. Thus, the matched-benchmark gains translate to higher overall reasoning accuracy under selective intervention without retuning the policy on the natural evaluation distributions.

\subsection{Ablation and Repair-Space Analysis}
\label{sec:exp_ablation}

\begin{wraptable}{r}{0.45\columnwidth}
\vspace{-22pt}
\centering
\scriptsize
\setlength{\tabcolsep}{2.5pt}
\caption{\textbf{Component ablations on DiffusionGemma + MetaMathQA.} Full \textsc{LOCKR}: $67.9\%$ accuracy at $3.1\times$ cost.}
\label{tab:lockr_ablation}
\begin{tabular*}{\linewidth}{@{\extracolsep{\fill}}lrrrr@{}}
\toprule
\textbf{Ablation} & \textbf{Acc. (\%)} & \textbf{Cost} & \textbf{$\Delta$Acc. (pp)} & \textbf{$\Delta$Cost} \\
\midrule
Single hidden snapshot & 64.7 & 3.0 & -3.2 & -0.1 \\
w/o branch trajectory & 63.9 & 3.1 & -4.0 & 0.0 \\
w/o trajectory GRU & 63.7 & 2.9 & -4.2 & -0.2 \\
w/o targeted remasking & 62.7 & 2.9 & -5.2 & -0.2 \\
w/o adaptive trigger & 67.8 & 5.0 & -0.1 & +1.9 \\
Random trigger & 59.1 & 3.1 & -8.8 & 0.0 \\
\bottomrule
\end{tabular*}
\vspace{-12pt}
\end{wraptable}

We finally examine which components drive end-to-end performance. Table~\ref{tab:lockr_ablation} summarizes the main component ablations on DiffusionGemma + MetaMathQA. Relative to full \textsc{LOCKR} ($67.9\%$ at $3.1\times$ cost), replacing the trajectory with a single snapshot or removing branch trajectories, GRU modeling, or targeted remasking reduces accuracy by $3.2$--$5.2$ percentage points. The trigger serves a different role: repairing every example preserves nearly the same accuracy but raises cost to $5.0\times$, whereas a random trigger at comparable cost loses $8.8$ percentage points. Thus, risk detection determines where computation is allocated, while targeted repair and trajectory-aware verification determine whether that computation is useful.

Oracle results further show that strong detection alone does not guarantee a large end-to-end gain. The smaller repair-oracle headroom on LLaDA-2 + Orca-Math ($65.9\%$ versus $71.7\%$ on MetaMathQA) indicates that performance is bounded both by whether the fixed bank exposes a correct continuation and by whether the selector ranks it first. Detailed repair-coverage, selection-gap, and cross-model ablations are provided in Appendices~\ref{app:repair_coverage} and~\ref{app:ablations}.

\noindent\textbf{Limitations.} \textsc{LOCKR} uses a fixed repair-action bank and is evaluated on mathematical reasoning with two white-box dLLM families. This creates a repairability ceiling: even strong failure detection cannot recover an example when the available action bank fails to expose a correct continuation, and correct branches may still be misranked when they are available. Broader domains, architectures, adaptive repair-space construction, and stronger trajectory-aware verification therefore remain important directions for future work. Our normalized cost measures denoising work rather than hardware-specific wall-clock latency, which depends on batching and serving infrastructure.

\section{Conclusion}

We identify SBW lock-in as a recurring dLLM reasoning failure and show, under surface-matched evaluation, that hidden-state trajectories provide stronger correctness signals than observable decoding statistics or single hidden snapshots. \textsc{LOCKR} turns these trajectories into a test-time planning signal that allocates repair computation, expands targeted counterfactual branches, and guides branch selection. Across two dLLMs and three reasoning benchmarks, this yields consistent gains in detection, branch selection, and end-to-end accuracy with selective intervention.

Overall, dLLM hidden trajectories serve not only as diagnostic traces, but as actionable state representations for deciding when and how to intervene during reasoning.

\section*{Reproducibility Statement}

We provide detailed definitions of SBW lock-in, the surface-matching protocol, hidden-state trajectory construction, repair actions, and branch-selection procedure in the main text and appendix. The appendix further specifies model and dataset preprocessing, generation settings, trajectory extraction, training and validation procedures, hyperparameters, and compute accounting. All reported comparisons use fixed data splits and validation-selected operating points, with test sets reserved for final evaluation. We also provide the implementation and experiment configurations through the anonymized code repository and supplementary materials to facilitate reproduction of the reported results.

\section*{AI Use Statement}

We used AI assistants for language editing, organization and refinement of manuscript text, LaTeX and code checking, and assistance in identifying implementation or experimental-pipeline errors. All research questions, conceptual contributions, methodological decisions, experimental design, implementation choices, execution of experiments, analysis of results, and scientific conclusions were determined and verified by the authors.

\bibliography{iclr2027_conference}
\bibliographystyle{iclr2027_conference}

\appendix

\section{Additional Experimental Details}
\label{app:experimental_details}

This appendix provides additional details of the experimental protocol summarized in Section~\ref{sec:exp_setup}. We describe the model and decoding configurations, dataset preprocessing and answer evaluation, data splitting and model-selection procedures, and the accounting used for relative inference cost. Details specific to surface matching, hidden-trajectory construction, and the repair planner are provided separately in Appendices~\ref{app:surface_features}, \ref{app:hidden_trajectory}, and \ref{app:repair_details}, respectively.

\subsection{Models and Generation Settings}
\label{app:model_generation}

\paragraph{DiffusionGemma.}
Our primary model is \texttt{google/diffusiongemma-26B-A4B-it} \citep{diffusiongemma2026}. We run the model in bfloat16 precision and use the Hugging Face block-diffusion generation implementation. Standard generations use a maximum output length of 256 tokens and 32 denoising steps. Unless otherwise stated, generation is deterministic, with temperature set to $0$. We record the decoded trajectory at every denoising step and evaluate early lock-in at the midpoint of the recorded trajectory, corresponding to a checkpoint fraction of $0.50$. The lock-in window uses the most recent half of the prefix observed by this checkpoint, as defined in Section~\ref{sec:early_lockin}. DiffusionGemma's native generation API determines its internal block-diffusion schedule; although our unified runner retains a \texttt{block\_length} argument for compatibility with the LLaDA-2 pipeline, this argument is not used to override the native DiffusionGemma generation procedure.

\paragraph{LLaDA-2.}
Our second model is \texttt{inclusionAI/LLaDA2.0-mini} \citep{bie2025llada20}. We likewise use bfloat16 precision, a maximum generation length of 256 tokens, and 32 denoising steps. LLaDA-2 generation uses a block length of 32 tokens. Base generations use temperature $0$, with no nucleus or top-$k$ sampling, and the decoded trajectory is recorded at every denoising step. The same midpoint checkpoint fraction of $0.50$ and the same early-lock-in definition are used as for DiffusionGemma. This shared outer protocol ensures that the operational definition of SBW lock-in is consistent across the two model families even though their underlying diffusion implementations differ.

\paragraph{Deterministic seeding.}
We use base seed 17 throughout the main experimental pipeline. For generation and replay operations that require sample-level randomness, each example receives a deterministic seed derived from the base seed and a stable sample key containing its split, sample identifier, and dataset index. This avoids dependence on Python's process-randomized hash function and makes sample trajectories invariant to multi-GPU sharding or worker assignment. The same stored sample seed is reused when hidden trajectories or repair starting states must be reconstructed from an earlier generation.

Table~\ref{tab:app_generation_settings} summarizes the common base-generation configuration. Sampling used specifically for repair branches is described separately in Appendix~\ref{app:repair_details}.

\begin{table}[t]
\centering
\small
\setlength{\tabcolsep}{3pt}
\caption{Base-generation settings used for SBW analysis. The DiffusionGemma implementation uses its native block-diffusion generation schedule; the \texttt{block\_length} compatibility argument does not override that schedule.}
\label{tab:app_generation_settings}
\begin{tabular}{p{0.31\columnwidth}p{0.30\columnwidth}p{0.27\columnwidth}}
\toprule
\textbf{Parameter} & \textbf{DiffusionGemma} & \textbf{LLaDA-2} \\
\midrule
Model checkpoint & \shortstack[l]{\texttt{diffusiongemma-}\\\texttt{26B-A4B-it}} & \shortstack[l]{\texttt{LLaDA2.0-}\\\texttt{mini}} \\
Precision & bfloat16 & bfloat16 \\
Maximum generation length & 256 & 256 \\
Denoising steps & 32 & 32 \\
Block length & native model schedule & 32 \\
Base temperature & 0.0 & 0.0 \\
Base top-$p$ / top-$k$ & not used & not used \\
Trajectory recording interval & every step & every step \\
Early-lock-in checkpoint fraction & 0.50 & 0.50 \\
Lock-in prefix fraction & 0.50 & 0.50 \\
Answer-region token budget & 8 & 8 \\
Base seed & 17 & 17 \\
\bottomrule
\end{tabular}
\end{table}

\subsection{Dataset Processing and Answer Evaluation}
\label{app:data_eval}

\paragraph{MetaMathQA.}
We construct the MetaMathQA evaluation pool from the numeric-answer portion of MetaMathQA \citep{yu2024metamath}. Our processed asset contains approximately 100K examples with a question, a reference solution, and a canonical numeric gold answer. We retain examples for which a deterministic numeric target can be extracted and represented in the common evaluation schema. The same processed pool is used across model families so that differences between DiffusionGemma and LLaDA-2 are not attributable to dataset preprocessing.

\paragraph{Orca-Math.}
The Orca-Math pool is derived from Orca-Math \citep{mitra2024orcamath} and stored as a 100K numeric-clean evaluation asset. During construction, questions are deduplicated, the candidate pool is shuffled using seed 17, and examples are filtered to retain instances with a usable and internally consistent numeric target. The final pool is capped at 100K examples. The resulting dataset is shared by the DiffusionGemma and LLaDA-2 evaluation pipelines.

\paragraph{NuminaMath V2.}
For NuminaMath \citep{numinamath2024}, we use a 100K numeric-clean V2 subset. To reduce contamination between the three mathematical reasoning pools used in our study, this version excludes examples assigned to the MetaMathQA and Orca-Math overlap sets during preprocessing. As with the other benchmarks, retained examples must admit a deterministic numeric gold answer under the common evaluation procedure. DiffusionGemma results in the main paper use this V2 pool. The same processed asset is also used for the LLaDA-2 preprocessing and surface-matching pipeline. For LLaDA-2, downstream repair evaluation is reported on MetaMathQA and Orca-Math; NuminaMath V2 is used only for benchmark-construction diagnostics.

\paragraph{Prompt and answer format.}
Across datasets, we use an answer-first generation format that encourages the model to expose a machine-readable numeric answer while still allowing subsequent reasoning text. The expected response begins with a final-answer marker of the form
\[
\texttt{\#\#\#\# <final numeric answer>}
\]
followed by optional reasoning. The same answer-extraction routine is applied to intermediate drafts and final generations, which is essential because early-lock-in labels are defined from the sequence of extracted answers rather than from the final output alone.

\paragraph{Numeric normalization.}
Answer evaluation is deterministic and does not rely on an auxiliary language-model judge. Before comparison, we remove common LaTeX wrappers such as \verb|\boxed{...}|, \verb|\fbox{...}|, and simple \verb|\text{...}| wrappers, strip formatting characters and thousands separators, and normalize simple fractional expressions. We then extract the final numeric token from the normalized answer string. Integer, decimal, and fractional representations are converted to a canonical numeric form whenever possible. For example, numerically equivalent forms such as \texttt{1/2}, \texttt{0.5}, and \texttt{0.500} are treated identically.

Let $\hat{a}$ and $g$ denote the normalized predicted and gold numeric values. We mark the prediction as correct when
\[
|\hat{a}-g|
\leq
\max\left(10^{-6},\,10^{-6}\max\left(|\hat{a}|,|g|\right)\right),
\]
which is equivalent to using absolute and relative tolerances of $10^{-6}$ under the standard \texttt{isclose} convention. If either answer cannot be normalized to a valid finite numeric value, the comparison is marked incorrect. The same correctness function is used for final-answer accuracy, early-lock-in labeling, repair-branch correctness, and repair-oracle analyses. This shared deterministic evaluator prevents changes in evaluation semantics across stages of the pipeline.

\subsection{Training and Validation Protocol}
\label{app:training_validation}

\paragraph{Matched-pair splitting for wrong-lock-in detection.}
The surface-matched benchmark consists of one early-correct and one early-wrong example per matched pair. We therefore split data at the \emph{pair} level rather than the individual-row level: both members of a matched pair are always assigned to the same partition. The first-stage wrong-lock-in detector uses a $70/10/20$ train/validation/test partition with seed 17. We first hold out $20\%$ of matched pairs for testing and then reserve $10\%$ of the complete dataset for validation. Stratified splitting is used when feasible, with deterministic seeded random splitting as a fallback. Feature normalization statistics are estimated from the training partition only and then applied unchanged to validation and test data.

The exact sample and pair identifiers assigned to the train, validation, train--validation, and test partitions are materialized and reused by downstream stages. In particular, repair branches are generated according to these frozen identifiers rather than by independently resplitting the data. This ensures that the test examples used for wrong-lock-in detection remain test examples throughout subsequent repair experiments.

\paragraph{Single hidden snapshot baselines.}
The single hidden snapshot baseline is deliberately given access to multiple candidate observation positions. Separate models are trained from hidden representations at relative prefix positions $\{0.25,0.50,0.75,1.00\}$. The snapshot used for final reporting is selected solely according to validation performance, after which the chosen model and snapshot position are evaluated once on the frozen test partition. Test performance is never used to choose the observation point.

\paragraph{Branch-verifier splitting.}
The second-stage branch verifier requires stronger grouping constraints because every original generation can produce multiple correlated repair branches. We therefore use a strict global grouping procedure that couples an original matched pair with all repair examples derived from either member of that pair. A repair branch can consequently never appear in training or validation when its corresponding original trajectory is assigned to the test partition. In the formal multi-task branch-verifier configuration, $20\%$ of these globally grouped units are reserved for testing and $15\%$ for validation, using seed 17. The remaining groups are used for training. This split is constructed before branch-verifier model selection and remains fixed for the downstream selector.

\paragraph{Model selection and operating points.}
All learned models, hyperparameters, early-stopping decisions, snapshot choices, and intervention operating points are selected without access to test outcomes. The first-stage trajectory detector predicts early-lock-in correctness, and wrong-lock-in risk is defined as one minus this score. Candidate intervention operating points are enumerated on the validation partition. We select the candidate with the highest validation end-to-end planner accuracy; exact accuracy ties are broken in favor of lower normalized inference cost. The chosen operating point is then frozen and applied unchanged to the held-out test set.

On the natural distribution, routing first applies the early-lock-in eligibility criterion using only the observed answer trajectory. Examples that are not early locked keep the baseline continuation and bypass both risk scoring and repair. Early-locked examples are assigned a wrong-lock-in risk, and the fixed repair bank is executed only when the frozen operating point triggers intervention. Natural-distribution repair rate is therefore defined globally as the number of examples actually sent to repair divided by the total number of natural-distribution examples; it is not conditioned on early-lock-in eligibility and is not tuned to a target test-set budget.

The branch trajectory model and the final selector are likewise trained using training data and selected using validation performance. The final test candidate pools are held fixed across branch-selection baselines so that comparisons among random selection, confidence, Surface-18, external PRM, single hidden snapshot representations, and \textsc{LOCKR} differ only in how the same available candidates are ranked. Oracle branch selection uses test labels only to quantify the fixed-pool selection upper bound. The end-to-end repair oracle additionally retains correct baseline outputs and rescues baseline-wrong outputs whenever a correct repair is present. Both oracle quantities are analysis-only and are never used for training or model selection.

\paragraph{Baseline implementation details.}
The external PRM baseline uses \texttt{Qwen2.5-Math-PRM-7B}. For each candidate, we provide the problem together with the complete candidate branch, use the model's official final reward as the branch score, and select the highest-scoring branch from the same fixed repair pool used by the other selectors. No checkpoint, reward aggregation rule, or threshold is selected using test outcomes. ReMDM is run with its official default inference-time hyperparameters and remasking rule, adapting only the underlying base model, prompt, and common 256-token decoding setting; it does not use \textsc{LOCKR} hidden states, the \textsc{LOCKR} risk detector, or the trajectory-aware branch verifier.

The Surface-18 LR and HGB detectors use only the same 18 observable decoding statistics used for matching, with standard scikit-learn implementations and preprocessing fit on training data only. The Surface-18 branch verifier likewise uses only surface features computed for candidates from the common fixed repair pool and selects the highest-scoring candidate within each pool. The Surface-18 planner is the surface-only counterpart to \textsc{LOCKR}: it uses a Surface-18 risk trigger, the same eight-action repair bank, and a Surface-18 branch selector. Its intervention operating point is chosen by the same validation objective used for \textsc{LOCKR}---maximum end-to-end validation accuracy, with lower normalized inference cost breaking ties---and is frozen before test evaluation.

\subsection{Compute and Relative-Cost Accounting}
\label{app:cost_accounting}

We report a normalized inference-cost measure to compare selective repair with methods that allocate different amounts of test-time computation. The measure is intended as a model-execution proxy rather than as a hardware-specific wall-clock metric. One standard generation under the corresponding model and dataset configuration is defined to have relative cost $1.0\times$.

Let $D_0$ denote the denoising work required by one standard base generation and let $D_i$ denote the total model-generation work actually executed for test example $i$, including the original trajectory and any additional repair or restart trajectories invoked by the evaluated policy. For a test set of $N$ examples, we report
\[
C_{\mathrm{rel}}
=
\frac{1}{N}
\sum_{i=1}^{N}
\frac{D_i}{D_0}.
\]
Thus, an example on which a selective method does not trigger repair contributes exactly one base-generation unit. When repair is triggered from an intermediate draft state, only the additional continuation work executed by the repair procedure is charged beyond the base trajectory; the reused prefix is not counted as a second full generation. In contrast, a method that performs an independent full restart is charged for the complete additional denoising trajectory.

For methods that generate several candidates, the costs of all candidates actually produced at inference time are accumulated before normalization. Consequently, always-repair and Best-of-$N$-style methods incur their additional cost on every test example, whereas \textsc{LOCKR} incurs repair cost only on examples selected by the first-stage risk policy. This distinction is important for the natural-distribution experiments, where only $22$--$41\%$ of examples are repaired in the five settings reported in the main paper.

The reported relative cost tracks diffusion-model denoising work. It excludes one-time offline training, dataset preprocessing, surface matching, and construction of training caches, and it also excludes auxiliary non-diffusion scoring-model compute such as forward passes through the external PRM. This convention is favorable to PRM-guided baselines: their reported relative costs reflect candidate-generation work but not the additional PRM scoring pass. We apply the same denoising-work accounting convention consistently within each comparison table.

Relative cost should therefore be interpreted as normalized diffusion-model denoising work rather than exact floating-point operations or elapsed time. Actual wall-clock latency depends on model architecture, hardware, batching, parallel execution of repair branches, and implementation details. The normalized measure is used because it permits controlled comparison of how efficiently different policies allocate additional denoising computation while remaining independent of a particular serving configuration.

\section{SBW Benchmark Construction}
\label{app:surface_features}

This appendix provides the full construction details for the surface-matched SBW benchmark introduced in Section~\ref{sec:surface_matching}. The objective of the matching procedure is to reduce observable differences between early-correct and early-wrong lock-in examples before any hidden-state information is used. We first define the 18 prefix-level surface statistics used for matching, then describe the exact normalization, caliper, distance, and one-to-one matching procedure, and finally report additional balance diagnostics.

\begin{table*}[t]
\centering
\small
\setlength{\tabcolsep}{3pt}
\caption{SBW prevalence in the five formal source populations. \textbf{Processed} denotes the formal surface-generation population used for checkpoint-local lock-in labeling; rows are not additionally filtered by whether a final answer can later be extracted. SBW rate is the early-wrong count divided by the processed count.}
\label{tab:app_sbw_prevalence}
\begin{tabular*}{\textwidth}{@{\extracolsep{\fill}}llrrrrr@{}}
\toprule
\textbf{Model} &
\textbf{Dataset} &
\textbf{Processed} &
\shortstack{\textbf{Early-}\\\textbf{correct}} &
\shortstack{\textbf{Early-}\\\textbf{wrong}} &
\shortstack{\textbf{Not early}\\\textbf{locked}} &
\shortstack{\textbf{SBW}\\\textbf{rate}} \\
\midrule
DiffusionGemma & MetaMathQA & 99,839 & 63,023 & 14,863 & 21,953 & 14.89\% \\
DiffusionGemma & Orca-Math & 100,000 & 52,495 & 20,568 & 26,937 & 20.57\% \\
DiffusionGemma & NuminaMath V2 & 100,000 & 28,264 & 36,217 & 35,519 & 36.22\% \\
LLaDA-2 & MetaMathQA & 99,839 & 37,855 & 59,688 & 2,296 & 59.78\% \\
LLaDA-2 & Orca-Math & 100,000 & 38,326 & 60,474 & 1,200 & 60.47\% \\
\bottomrule
\end{tabular*}
\end{table*}

\subsection{Full Surface-18 Feature Definitions}
\label{app:surface18_def}

All surface features are computed using information available no later than the early-lock-in checkpoint. Let the surface recorder produce $M$ valid update records within this prefix. For update $r\in\{1,\ldots,M\}$, let $q_r$ denote the confidence assigned to the selected token, let $e_r$ denote the entropy of the corresponding token distribution, and let $m_r$ denote the probability margin between the highest- and second-highest-probability tokens. Given a token distribution $\mathbf{p}_r$, these quantities are
\[
e_r=-\sum_v p_{r,v}\log p_{r,v},
\qquad
m_r=p_{r,(1)}-p_{r,(2)},
\]
where $p_{r,(1)}$ and $p_{r,(2)}$ are the largest and second-largest probabilities, respectively. We additionally define
\[
\ell_r=\log\left(\max(q_r,10^{-12})\right).
\]

The surface recorder tracks these statistics for the generation positions associated with the answer-focused region while the prefix is being denoised. Only records whose global denoising step does not exceed the early-lock-in checkpoint are retained. Examples for which any of the 18 required matching features is missing or non-finite are excluded before matching.

For a scalar sequence $\{y_r\}_{r=1}^{M}$, we use the population standard deviation
\[
\operatorname{Std}(y)=
\sqrt{\frac{1}{M}\sum_{r=1}^{M}(y_r-\bar{y})^2}.
\]
We also use a normalized trapezoidal area under the trajectory,
\[
\operatorname{AUC}(y)=
\frac{1}{M-1}
\sum_{r=1}^{M-1}
\frac{y_r+y_{r+1}}{2},
\]
for $M>1$; when $M=1$, the AUC is defined as the single observed value.

Table~\ref{tab:surface18_definitions} lists the complete Surface-18 representation.

\begin{table*}[t]
\centering
\small
\setlength{\tabcolsep}{4pt}
\caption{Complete definitions of the 18 prefix-level observable features used for surface matching. All quantities are computed using records available no later than the early-lock-in checkpoint.}
\label{tab:surface18_definitions}
\begin{tabular}{p{0.32\textwidth}p{0.62\textwidth}}
\toprule
\textbf{Feature} & \textbf{Definition} \\
\midrule
\texttt{mean\_confidence} & Mean selected-token confidence, $\frac{1}{M}\sum_r q_r$. \\
\texttt{mean\_entropy} & Mean token-distribution entropy, $\frac{1}{M}\sum_r e_r$. \\
\texttt{mean\_margin} & Mean top-1 versus top-2 probability margin, $\frac{1}{M}\sum_r m_r$. \\
\texttt{min\_confidence} & Minimum selected-token confidence, $\min_r q_r$. \\
\texttt{min\_margin} & Minimum top-1 versus top-2 probability margin, $\min_r m_r$. \\
\texttt{token\_flip\_count} & Total number of changes in the predicted token at tracked generation positions across successive observations before the checkpoint. Changes are counted independently at each position and then summed. \\
\texttt{answer\_string\_flip\_count} & Number of transitions between different normalized, non-empty extracted answers in the recorded answer trajectory up to the checkpoint. Empty answer observations are ignored when counting transitions. \\
\texttt{first\_answer\_index} & Zero-based index of the first recorded denoising state at which a valid non-empty normalized answer can be extracted. \\
\texttt{mean\_logprob} & Mean log confidence, $\frac{1}{M}\sum_r \ell_r$. \\
\texttt{sum\_logprob} & Cumulative log confidence, $\sum_r \ell_r$. \\
\texttt{min\_logprob} & Minimum log confidence, $\min_r \ell_r$. \\
\texttt{std\_confidence} & Population standard deviation of $\{q_r\}_{r=1}^{M}$. \\
\texttt{std\_entropy} & Population standard deviation of $\{e_r\}_{r=1}^{M}$. \\
\texttt{std\_margin} & Population standard deviation of $\{m_r\}_{r=1}^{M}$. \\
\texttt{entropy\_AUC} & Normalized trapezoidal AUC of the ordered entropy sequence $\{e_r\}_{r=1}^{M}$. \\
\texttt{confidence\_AUC} & Normalized trapezoidal AUC of the ordered confidence sequence $\{q_r\}_{r=1}^{M}$. \\
\texttt{margin\_AUC} & Normalized trapezoidal AUC of the ordered margin sequence $\{m_r\}_{r=1}^{M}$. \\
\texttt{path\_entropy} & Cumulative prefix entropy, $\sum_r e_r$. The implementation retains the name \texttt{path\_entropy}; it is the accumulated entropy over observed surface updates rather than the entropy of the discrete answer-string path. \\
\bottomrule
\end{tabular}
\end{table*}

These features intentionally cover several complementary aspects of observable convergence. The confidence, entropy, and margin statistics capture instantaneous model uncertainty; their minima and standard deviations capture local weakness and dispersion; the AUC features summarize their evolution through the observed prefix; token and answer flips quantify surface instability; first-answer timing captures answer emergence; and the log-probability statistics and cumulative entropy summarize confidence accumulated over the trajectory. Importantly, none of these features uses hidden activations.

\subsection{Matching Algorithm and Calipers}
\label{app:matching_algorithm}

\paragraph{Candidate pools.}
For each model--dataset setting, we use the operational early-lock-in labels defined in Section~\ref{sec:early_lockin}. The positive candidate pool consists of \texttt{early\_correct\_lockin} examples and the negative candidate pool consists of \texttt{early\_wrong\_lockin} examples. Samples outside these two categories are not considered for matching. We use the stable sample identifier as the primary unique key rather than shard-local indices, since local indices may be repeated after multi-worker generation outputs are merged. Samples with a missing or non-finite value in any Surface-18 dimension are removed before candidate construction.

\paragraph{Robust normalization.}
Let $s_{ij}$ denote feature $j$ for candidate example $i$. For each model--dataset setting, the early-correct and early-wrong candidate pools are combined before computing the matching transformation. For feature $j$, let $\widetilde{s}_j$ denote the median of the combined pool and let
\[
\operatorname{IQR}_j=Q_{0.75,j}-Q_{0.25,j}.
\]
We robustly scale each feature as
\[
z_{ij}=
\frac{s_{ij}-\widetilde{s}_j}
{\operatorname{IQR}_j}.
\]
If $\operatorname{IQR}_j<10^{-8}$, we set the denominator to $1$ to avoid numerical instability. This transformation is used only to define matching geometry; downstream predictive models use their own training-only preprocessing as described in Appendix~\ref{app:training_validation}.

\paragraph{Caliper constraints.}
We use the same \emph{refined} matching preset for all formal model--dataset settings. Candidate pairs must first satisfy feature-specific calipers. Two discrete answer-trajectory statistics use calipers in their original scale: \texttt{answer\_string\_flip\_count} must match exactly, while \texttt{first\_answer\_index} may differ by at most two recorded answer states. The remaining constrained features use absolute differences in robust-scaled feature space. Table~\ref{tab:matching_calipers} gives the exact calipers and matching weights.

\begin{table*}[t]
\centering
\scriptsize
\caption{Feature weights and calipers for the formal \emph{refined} Surface-18 matcher. A dash indicates that no additional caliper of that type is imposed. All 18 features remain part of the weighted distance even when only a raw-scale caliper is used.}
\label{tab:matching_calipers}
\begin{tabular}{lccc}
\toprule
\textbf{Feature} & \textbf{Distance weight $w_j$} & \textbf{Scaled caliper} & \textbf{Raw-scale caliper} \\
\midrule
\texttt{mean\_confidence} & 1.50 & 0.75 & -- \\
\texttt{mean\_entropy} & 1.50 & 0.75 & -- \\
\texttt{mean\_margin} & 1.50 & 0.75 & -- \\
\texttt{min\_confidence} & 1.00 & 1.00 & -- \\
\texttt{min\_margin} & 1.00 & 1.00 & -- \\
\texttt{token\_flip\_count} & 1.25 & 1.00 & -- \\
\texttt{answer\_string\_flip\_count} & 2.00 & -- & 0 \\
\texttt{first\_answer\_index} & 1.50 & -- & 2 \\
\texttt{mean\_logprob} & 1.25 & 0.75 & -- \\
\texttt{sum\_logprob} & 1.00 & 1.25 & -- \\
\texttt{min\_logprob} & 1.00 & 1.00 & -- \\
\texttt{std\_confidence} & 1.00 & 1.00 & -- \\
\texttt{std\_entropy} & 1.00 & 1.00 & -- \\
\texttt{std\_margin} & 1.00 & 1.00 & -- \\
\texttt{entropy\_AUC} & 1.25 & 0.75 & -- \\
\texttt{confidence\_AUC} & 1.25 & 0.75 & -- \\
\texttt{margin\_AUC} & 1.25 & 0.75 & -- \\
\texttt{path\_entropy} & 1.00 & 1.25 & -- \\
\bottomrule
\end{tabular}
\end{table*}

\paragraph{Weighted matching distance.}
For an early-correct candidate $i$ and an early-wrong candidate $k$ that satisfy all applicable calipers, we define the matching distance
\[
d(i,k)=
\sqrt{
\sum_{j=1}^{18}
w_j\left(z_{ij}-z_{kj}\right)^2
},
\]
where $w_j$ is the feature-specific weight in Table~\ref{tab:matching_calipers}. We retain only candidate edges satisfying
\[
d(i,k)\leq 3.50.
\]
For computational efficiency and to prevent very dense candidate regions from dominating the edge set, at most the 160 lowest-distance early-wrong candidates are retained for each early-correct anchor after applying the calipers and distance cap.

\paragraph{One-to-one greedy selection.}
All retained candidate edges are pooled and sorted in ascending order of weighted distance. We then greedily accept the lowest-distance edge whose early-correct and early-wrong examples have not previously been selected. Matching is therefore performed without replacement: every source example can appear in at most one matched pair. Seed 17 is used as a deterministic random tie-breaker between edges with identical or effectively indistinguishable distances. Formal runs permit at most 40,000 pairs, although the available caliper-constrained pools determine the smaller realized pair counts reported below.

This procedure differs from learning a classifier that directly optimizes group separation. Matching uses only observable Surface-18 features and the early-correct/early-wrong group labels required to construct the controlled benchmark; hidden representations are not extracted or consulted until after the matched pairs have been frozen.

\subsection{Additional Balance Statistics}
\label{app:balance_stats}

Figure~\ref{fig:surface_balance} visualizes the before--after balance shift for the representative DiffusionGemma + MetaMathQA setting. Here we report the corresponding post-matching diagnostics for all constructed benchmark pools. We evaluate post-matching balance independently for every Surface-18 feature. For a feature $j$, let $\mu_j^{+}$ and $\sigma_j^{+}$ denote its sample mean and sample standard deviation among matched early-correct examples and let $\mu_j^{-}$ and $\sigma_j^{-}$ denote the corresponding quantities among matched early-wrong examples. We use
\[
\operatorname{SMD}_j=
\frac{\mu_j^{+}-\mu_j^{-}}
{\sqrt{\left((\sigma_j^{+})^2+(\sigma_j^{-})^2\right)/2}},
\]
where the within-group standard deviations are computed with the sample standard-deviation convention. If the denominator is numerically zero, the SMD is set to zero. We report both the signed and absolute SMD for every feature, together with the group means, medians, standard deviations, and the 10th, 25th, 50th, 75th, and 90th percentiles in the machine-readable matching summaries.

Table~\ref{tab:app_matching_balance} summarizes balance across all surface benchmarks constructed with the formal protocol. For completeness, this table includes the LLaDA-2 + NuminaMath V2 surface benchmark, whose matching stage was completed independently of the downstream repair experiments. Across all six constructed model--dataset pools, the mean absolute SMD is at most $0.01765$ and the maximum absolute SMD is at most $0.08494$.

\begin{table*}[t]
\centering
\small
\caption{Surface-matching balance across the six constructed model--dataset benchmark pools. The final column reports the largest absolute SMD among the 18 matched surface features. LLaDA-2 + NuminaMath V2 is included here as a completed benchmark-construction setting even when downstream experiments are reported separately.}
\label{tab:app_matching_balance}
\begin{tabular}{llrrr}
\toprule
\textbf{Model} & \textbf{Dataset} & \textbf{Matched pairs} & \textbf{Mean $|\mathrm{SMD}|$} & \textbf{Max $|\mathrm{SMD}|$} \\
\midrule
DiffusionGemma & MetaMathQA & 13,376 & 0.01629 & 0.06530 \\
DiffusionGemma & Orca-Math & 16,067 & 0.01765 & 0.08035 \\
DiffusionGemma & NuminaMath V2 & 19,137 & 0.01622 & 0.08494 \\
LLaDA-2 & MetaMathQA & 25,064 & 0.00947 & 0.03416 \\
LLaDA-2 & Orca-Math & 22,418 & 0.01186 & 0.04639 \\
LLaDA-2 & NuminaMath V2 & 19,468 & 0.00988 & 0.03483 \\
\bottomrule
\end{tabular}
\end{table*}

Across all constructed settings, the maximum absolute SMD remains below $0.085$, indicating close post-matching balance across the 18 controlled surface dimensions. Balance is not driven by only a small subset of features: the reported maximum is taken over all 18 dimensions, so even the least-balanced matched feature remains below $0.085$ in every constructed setting. For example, on LLaDA-2 + Orca-Math, the largest remaining difference is observed for \texttt{entropy\_AUC}, with $|\mathrm{SMD}|=0.0464$, while exact matching on \texttt{answer\_string\_flip\_count} yields an SMD of zero for that feature.

These diagnostics do not imply that the two groups are identical in every observable aspect. Rather, they substantially reduce the particular confidence, uncertainty, stability, emergence, and likelihood differences represented by Surface-18. The resulting benchmark is therefore designed to test a more specific question than ordinary correctness prediction: whether internal denoising representations contain information about erroneous lock-in after a broad set of readily observable prefix statistics has been controlled.

\section{Hidden-Trajectory Representation}
\label{app:hidden_trajectory}

This appendix provides the implementation details of the hidden-trajectory representation used by \textsc{LOCKR}. The same representation geometry is used for prefix-level wrong-lock-in detection and, unless otherwise noted, for repair-branch trajectories. We first describe how hidden states are aligned to the denoising prefix, then define the fixed random-sign projection, specify the answer and change token tracks, and finally give the architecture and optimization details of the first-stage trajectory detector. The second-stage multi-task branch verifier is described separately in Appendix~\ref{app:repair_details}.

\subsection{Hidden-State Extraction}
\label{app:hidden_extraction}

\paragraph{Prefix restriction.}
For the first-stage detector, hidden representations are restricted to model states available no later than the early-lock-in checkpoint. Given the replayed or captured denoising trajectory for an example, we first identify the checkpoint used to assign its early-lock-in label and discard later hidden states for risk detection. Whenever available, we reuse the checkpoint index and global denoising step stored by the corresponding surface-generation record rather than recomputing the checkpoint independently. This keeps the hidden trajectory aligned with the exact prefix from which the surface-matched benchmark was constructed.

Let the resulting prefix contain $N_{\mathrm{pre}}$ hidden-state snapshots. We select four relative positions
\[
\mathcal{R}=\{0.25,0.50,0.75,1.00\}.
\]
For a relative position $r\in\mathcal{R}$, the selected prefix index is
\[
j(r)=
\operatorname{round}
\left(
r\left(N_{\mathrm{pre}}-1\right)
\right),
\]
clipped to the valid range $[0,N_{\mathrm{pre}}-1]$. Thus, the final selected snapshot corresponds to the endpoint of the observed prefix, while the earlier snapshots provide intermediate views of the same denoising process.

\paragraph{Layer selection.}
At every selected snapshot, we extract hidden representations from model layers
\[
\mathcal{L}=\{5,10,15,19\}.
\]
These four layers are fixed for all main experiments. Let
\[
\mathbf{h}_{r,l,p}\in\mathbb{R}^{d_h}
\]
denote the hidden representation at relative snapshot $r$, layer $l$, and token position $p$, where $d_h$ is the native hidden dimension of the corresponding model.

\paragraph{Model-specific capture.}
The two evaluated model families expose hidden states through slightly different generation implementations. For LLaDA-2, we deterministically replay the stored generation trajectory using the per-sample seed saved by the surface-generation stage, reconstruct the selected draft states, and execute a forward pass with hidden-state output enabled at each selected snapshot. For DiffusionGemma, hidden states are captured from the decoder canvas used during the model's native block-diffusion generation and aligned to the corresponding recorded denoising steps. In both cases, hidden extraction is tied to the same deterministic generation instance used to construct the surface record.

We perform consistency checks between replayed and stored generation metadata before retaining an example. These checks include the final extracted answer, the checkpoint answer, and the checkpoint location whenever the corresponding reference values are available. The purpose is to ensure that a hidden trajectory is not paired with a surface label produced by a different stochastic realization of the same prompt. The extracted feature artifacts store the selected denoising steps and replay diagnostics together with the compressed hidden representation.

\paragraph{Shared representation geometry.}
After snapshot and layer selection, both model families use the same representation layout:
\[
\underbrace{4}_{\text{prefix snapshots}}
\times
\underbrace{4}_{\text{layers}}
\times
\underbrace{2}_{\text{token tracks}}
\times
\underbrace{8}_{\text{positions per track}}
\times
\underbrace{128}_{\text{projected dimensions}}
=
32768.
\]
The resulting feature is stored as a 32,768-dimensional float vector. Its logical ordering is
\[
\text{snapshot}
\rightarrow
\text{layer}
\rightarrow
\text{track}
\rightarrow
\text{token position}
\rightarrow
\text{projected dimension}.
\]
No raw full-dimensional hidden activations are required by the downstream detector once this representation has been constructed.

\begin{table}[t]
\centering
\small
\caption{Hidden-trajectory representation used in the main experiments.}
\label{tab:hidden_representation_config}
\begin{tabular}{ll}
\toprule
\textbf{Component} & \textbf{Setting} \\
\midrule
Prefix-relative snapshots & $0.25, 0.50, 0.75, 1.00$ \\
Selected layers & $5, 10, 15, 19$ \\
Token tracks & answer region, changed answer region \\
Positions per track & 8 \\
Projection dimension & 128 \\
Projection type & fixed layer-specific random sign \\
Flattened feature dimension & 32,768 \\
GRU sequence length & 256 \\
\bottomrule
\end{tabular}
\end{table}

\subsection{Random-Sign Projection}
\label{app:random_projection}

The native hidden dimension of the underlying diffusion model is substantially larger than the representation width used by the trajectory detector. We therefore project every selected hidden vector to a fixed 128-dimensional space before trajectory modeling. The projection is random but fixed: it is generated once from a deterministic seed and is not learned jointly with the detector.

For each selected model layer $l$, we construct a layer-specific matrix
\[
\mathbf{R}_l\in\mathbb{R}^{d_h\times d_p},
\qquad
d_p=128,
\]
whose entries are independent Rademacher signs scaled by the projection width:
\[
(\mathbf{R}_l)_{ab}
=
\frac{\xi_{ab}}{\sqrt{d_p}},
\qquad
\xi_{ab}\in\{-1,+1\}.
\]
The projected representation is
\[
\mathbf{z}_{r,l,p}
=
\mathbf{h}_{r,l,p}\mathbf{R}_l
\in\mathbb{R}^{128}.
\]

Projection matrices are deterministic and layer specific. In the implementation, the random generator for layer $l$ is initialized from the run-level projection seed with a deterministic layer-dependent offset, so the same layer always uses the same matrix within a formal extraction run. The projection seed is fixed before detector training, stored in the feature-artifact metadata, and is never selected using validation or test performance. This also makes hidden features reproducible across sharded extraction workers.

The projection serves two practical purposes. First, it provides a common fixed-width representation across model families with different native hidden dimensions. Second, it allows the full multi-step, multi-layer trajectory to be stored and modeled without retaining the original high-dimensional activation tensors. Projection is performed independently at each selected token position before the trajectory is flattened or passed to the GRU.

\subsection{Answer and Change Tracks}
\label{app:hidden_tracks}

We retain two complementary token tracks at every selected denoising snapshot. Both tracks operate on the beginning of the generated sequence, consistent with the answer-first prompting format described in Appendix~\ref{app:data_eval}.

\paragraph{Answer-region track.}
The answer-region track uses the first
\[
K_{\mathrm{ans}}=8
\]
generation positions immediately following the prompt. If $p_0$ denotes the first generated-token position, the nominal answer positions are
\[
\mathcal{P}^{\mathrm{ans}}
=
\{p_0,p_0+1,\ldots,p_0+7\}.
\]
For each selected snapshot and layer, we extract the hidden state at these positions and apply the layer-specific projection defined above. If a position is not yet present in the active decoding window, its projected representation is replaced by a zero vector so that the feature shape remains fixed.

The answer-first prompt makes these early generated positions particularly relevant to the model's currently exposed answer. The track itself, however, does not depend on whether a token is numerically interpretable or whether the current extracted answer is correct.

\paragraph{Change track.}
The second track emphasizes positions that are actively changing across the selected trajectory. Let
\[
\mathbf{x}^{(r)}_{\mathrm{ans}}
=
\left(
x^{(r)}_{1},\ldots,x^{(r)}_{8}
\right)
\]
denote the token identities in the first eight generated positions at selected snapshot $r$. For two successive selected snapshots, we define the changed-position set
\[
\mathcal{C}^{(r)}
=
\left\{
k:
x^{(r)}_{k}
\neq
x^{(r^-)}_{k}
\right\},
\]
where $r^-$ denotes the preceding selected snapshot. At the first selected snapshot, the reference vector is initialized to mask tokens, so positions that have already departed from the masked state can enter the change track.

We retain up to
\[
K_{\mathrm{chg}}=8
\]
changed positions in their positional order. Their hidden states are projected using the same layer-specific projection matrix as the answer track. If fewer than eight positions change, the remaining slots are filled with zeros. The track therefore has a fixed shape even when the number of token changes varies across snapshots.

The two tracks capture different aspects of the internal trajectory. The answer track follows a stable spatial region throughout denoising, while the change track emphasizes locations undergoing representational and token-level revision. Both are retained because a SBW answer may exhibit little observable answer change even while internal representations at nearby or recently changing positions continue to evolve.

\paragraph{Feature assembly.}
For every selected snapshot $r$ and layer $l$, the projected answer and change tensors each have shape
\[
8\times128.
\]
We append the answer track followed by the change track for each layer, then proceed through the selected layers and snapshots in fixed order. The complete tensor can therefore be viewed conceptually as
\[
\mathbf{Z}
\in
\mathbb{R}^{4\times4\times2\times8\times128}.
\]
The stored representation is the flattened form
\[
\mathbf{H}
=
\operatorname{vec}(\mathbf{Z})
\in
\mathbb{R}^{32768}.
\]
For trajectory modeling, this vector is reshaped as a sequence of
\[
256
=
4\times4\times2\times8
\]
elements, each represented by a 128-dimensional projected hidden vector.

\subsection{GRU Architecture and Hyperparameters}
\label{app:gru_details}

The first-stage wrong-lock-in detector uses a separate bidirectional GRU trained only on original matched early-lock-in trajectories. It is distinct from the multi-task branch verifier described in Appendix~\ref{app:repair_details}.

\paragraph{Input normalization.}
For each of the 32,768 flattened feature dimensions, we compute the mean and standard deviation using training examples only. Each trajectory vector is standardized as
\[
\widetilde{\mathbf{H}}
=
\frac{\mathbf{H}-\boldsymbol{\mu}_{\mathrm{train}}}
{\boldsymbol{\sigma}_{\mathrm{train}}},
\]
where every standard-deviation entry is lower-bounded by $10^{-6}$ for numerical stability. The same training-derived statistics are then applied unchanged to validation and test examples.

The standardized vector is reshaped to
\[
\widetilde{\mathbf{H}}
\in
\mathbb{R}^{256\times128}.
\]
Each 128-dimensional element is first mapped through a learned linear projection
\[
\mathbf{u}_t
=
\mathbf{W}_{\mathrm{in}}\widetilde{\mathbf{h}}_t
+
\mathbf{b}_{\mathrm{in}},
\qquad
\mathbf{u}_t\in\mathbb{R}^{256}.
\]

\paragraph{Bidirectional GRU encoder.}
The projected sequence is processed by a single bidirectional GRU with hidden size 256 in each direction. Let $\overrightarrow{\mathbf{g}}$ and $\overleftarrow{\mathbf{g}}$ denote the final forward and backward hidden states. We concatenate them to obtain
\[
\mathbf{g}
=
\left[
\overrightarrow{\mathbf{g}};
\overleftarrow{\mathbf{g}}
\right]
\in\mathbb{R}^{512},
\]
and apply layer normalization before classification.

The prediction head is
\[
512
\rightarrow
256
\rightarrow
1,
\]
with dropout $0.30$ before each linear stage and a GELU nonlinearity after the first linear transformation. The resulting scalar logit is converted through a sigmoid to obtain
\[
f_{\theta}(\mathbf{H})
=
P_{\theta}
\left(
\text{early-correct lock-in}
\mid
\mathbf{H}
\right).
\]
As defined in Section~\ref{sec:lockr_risk}, the wrong-lock-in risk used by the planner is
\[
p_{\mathrm{risk}}
=
1-f_{\theta}(\mathbf{H}).
\]

\paragraph{Optimization.}
The detector is trained with binary cross-entropy on the early-correct label. Class weighting is computed automatically from the training split as the ratio of negative to positive examples; because the surface-matched benchmark is pair balanced, this weight is typically close to one. We optimize with AdamW using learning rate $10^{-4}$ and weight decay $3\times10^{-4}$. Gradients are clipped to a maximum norm of $1.0$.

Training runs for at most 40 epochs. The model checkpoint is selected by validation AUROC, with early stopping patience 7 and a minimum required improvement of $10^{-4}$. A ReduceLROnPlateau scheduler halves the learning rate after two validation plateaus. The standard training batch size is 256 and the evaluation batch size is 512; larger 512/1024 train/evaluation batches are used in settings where GPU memory permits, without changing the model or optimization objective. All detector training uses seed 17 and the pair-level train/validation/test partitions described in Appendix~\ref{app:training_validation}.

\begin{table}[t]
\centering
\small
\caption{First-stage trajectory-detector architecture and optimization settings.}
\label{tab:gru_hyperparameters}
\begin{tabular}{ll}
\toprule
\textbf{Parameter} & \textbf{Setting} \\
\midrule
Flattened input dimension & 32,768 \\
Sequence length & 256 \\
Projected element dimension & 128 \\
Learned input dimension & 256 \\
GRU type & bidirectional \\
GRU hidden size & 256 per direction \\
Pooled representation & 512 \\
Classifier hidden size & 256 \\
Dropout & 0.30 \\
Loss & binary cross-entropy with logits \\
Optimizer & AdamW \\
Learning rate & $1\times10^{-4}$ \\
Weight decay & $3\times10^{-4}$ \\
Gradient clipping & 1.0 \\
Maximum epochs & 40 \\
Early-stopping patience & 7 \\
Minimum improvement & $1\times10^{-4}$ \\
LR scheduler & ReduceLROnPlateau, factor $0.5$ \\
LR scheduler patience & 2 \\
Validation monitor & AUROC \\
Training seed & 17 \\
\bottomrule
\end{tabular}
\end{table}

This representation deliberately separates the extraction geometry from the downstream classifier. The selected layers, prefix-relative snapshots, token-track budgets, and random projection are fixed before detector training, while only the input projection, GRU encoder, LayerNorm parameters, and prediction head are optimized. This design also permits the single hidden snapshot baseline in Section~\ref{sec:exp_detection} to reuse the same underlying hidden features while removing temporal coverage, yielding a controlled comparison between static and trajectory-level internal representations.

\section{Repair and Planning Details}
\label{app:repair_details}

This appendix provides additional implementation details for the repair and branch-selection stages introduced in Sections~\ref{sec:lockr_repair} and~\ref{sec:lockr_verification}. We first specify the fixed repair action bank, then describe the model-specific targeted-remasking procedures, the multi-task branch-trajectory verifier, and the histogram gradient-boosted selector used to combine trajectory evidence with lightweight branch-level context.

\subsection{Repair Action Bank}
\label{app:repair_action_bank}

\textsc{LOCKR} uses a fixed repair action bank rather than learning repair locations or remasking budgets from training data. As defined in Section~\ref{sec:lockr_overview}, each action pairs an intervention checkpoint with a remasking budget and produces one counterfactual branch. We fix $B=8$ before training as a breadth--compute design choice that exposes several intervention times and strengths while keeping per-example branch expansion bounded. This bank is shared unchanged across all model--dataset settings and is not selected using validation or test performance. For every generation selected for intervention, we therefore construct exactly eight candidate branches. Each branch is associated with one checkpoint fraction and one remasking budget; the two lists are paired by index rather than combined as a Cartesian product.

Let
\[
\boldsymbol{\rho}
=
(0.10,0.20,0.25,0.33,0.40,0.50,0.66,0.80)
\]
denote the repair checkpoint fractions and let
\[
\mathbf{K}
=
(32,64,96,128,160,192,224,256)
\]
denote the corresponding nominal remasking budgets. Repair action $i$ is therefore
\[
\mathcal{A}_i=(\rho_i,K_i),
\qquad
i\in\{1,\ldots,8\}.
\]
The resulting action bank is listed in Table~\ref{tab:repair_action_bank}.

\begin{table}[t]
\centering
\small
\caption{Fixed repair action bank used in the main experiments. Checkpoint fractions and remasking budgets are paired by row; the bank therefore contains eight actions rather than $8\times8$ combinations.}
\label{tab:repair_action_bank}
\begin{tabular}{ccc}
\toprule
\textbf{Action} & \textbf{Checkpoint fraction} & \textbf{Nominal remask budget} \\
\midrule
1 & 0.10 & 32 \\
2 & 0.20 & 64 \\
3 & 0.25 & 96 \\
4 & 0.33 & 128 \\
5 & 0.40 & 160 \\
6 & 0.50 & 192 \\
7 & 0.66 & 224 \\
8 & 0.80 & 256 \\
\bottomrule
\end{tabular}
\end{table}

The repair checkpoint is selected from the original baseline trajectory. For a requested fraction $\rho_i$, the corresponding stored draft state is chosen from the recorded baseline denoising sequence and used as the starting canvas for action $\mathcal{A}_i$. This procedure is implemented as \texttt{repair\_checkpoint\_source=branch\_cycle}, while the continuation itself uses the checkpoint decoder-input path rather than restarting from the initial noisy sequence.

The repair bank intentionally spans both early and late baseline states. Because the first-stage risk decision is made at the early-lock-in checkpoint, actions with $\rho_i<0.50$ revisit states that are already part of the observed prefix, while actions with larger fractions may use baseline draft states reached after that checkpoint. These later states do not alter the original trigger decision: the first-stage wrong-lock-in risk remains fixed once computed from the prefix described in Section~\ref{sec:lockr_risk}. They serve only as candidate starting states for repair.

All repair branches use 96 continuation steps. Unlike deterministic base decoding, branch generation is stochastic in the main configuration. We use 96 continuation steps with temperature $1.0$, top-$p=0.95$, and top-$k=50$, with sampling enabled. A deterministic branch seed is associated with each sample--action pair so that the branch cache can later be exactly replayed for hidden-state extraction. Consequently, branch diversity comes from both the structured variation in checkpoint/remasking configuration and stochastic continuation, while the complete candidate pool remains reproducible once generated.

\begin{table}[t]
\centering
\small
\caption{Repair-generation settings used for the formal branch pools.}
\label{tab:repair_generation_config}
\begin{tabular}{ll}
\toprule
\textbf{Parameter} & \textbf{Setting} \\
\midrule
Number of repair actions & 8 \\
Repair checkpoint source & branch cycle \\
Branch mode & checkpoint decoder input \\
Repair continuation steps & 96 \\
Sampling & enabled \\
Temperature & 1.0 \\
Top-$p$ & 0.95 \\
Top-$k$ & 50 \\
Checkpoint fractions & $0.10$--$0.80$ \\
Nominal remask budgets & $32$--$256$ \\
\bottomrule
\end{tabular}
\end{table}

\subsection{Targeted Remasking Procedure}
\label{app:targeted_remasking}

The two model families expose different low-level continuation interfaces, so the exact mechanism used to construct a partially remasked checkpoint canvas is model specific. The high-level intervention principle is nevertheless shared: preserve most of the existing draft and reopen an answer-related portion of the generation for further denoising. Gold answers are never used to determine which positions are remasked.

\paragraph{DiffusionGemma.}
For DiffusionGemma, the formal branch generator uses the \texttt{answer\_or\_numeric} remasking policy on the checkpoint decoder canvas. The procedure first attempts to locate the token span corresponding to an answer already exposed by the baseline trajectory, using the available locked or replayed answer strings. Matching is preferentially performed in the later portion of the decoder canvas to reduce accidental matches to unrelated numbers appearing earlier in the reasoning trace.

If an answer span is located, we construct a localized window around this span and replace the selected token positions with the model's mask token. The nominal action budget $K_i$ controls the intervention scale, although the realized window may be expanded when necessary to cover the complete detected answer span together with a small amount of surrounding context.

If no reliable answer span can be located, the procedure falls back to a numeric search. We inspect the tail of the checkpoint canvas for tokens containing digits and identify the last approximately contiguous numeric group, which is likely to correspond to the exposed final answer in the answer-first generation format. A localized window around this numeric region is then remasked. If neither an answer span nor a numeric tail can be identified, the final fallback remasks an output suffix. This gives the formal fallback sequence
\[
\text{answer span}
\rightarrow
\text{numeric tail}
\rightarrow
\text{output suffix}.
\]

This hierarchy makes the intervention answer directed without requiring the answer extractor to succeed at every intermediate checkpoint. It also avoids using the gold answer to construct repair branches.

\paragraph{LLaDA-2.}
The LLaDA-2 continuation backend directly exposes localized remasking of generated tokens. Because the prompt format places the machine-readable answer at the beginning of the generated region, we remask the first $K_i$ generated positions available within the selected checkpoint window, clipped to the active sequence length. The selected positions are reset to the LLaDA-2 mask token and locally denoised during the repair continuation.

Thus, the precise remasking implementation differs between the two model families: DiffusionGemma explicitly localizes an exposed answer or numeric region on its decoder canvas, whereas the LLaDA-2 backend directly reopens an answer-prefix region. Both implementations satisfy the same intervention constraint used in the main text: the prompt and the majority of the existing draft are retained while an answer-related portion of the generation is made editable again.

\paragraph{Branch replay and correctness.}
For every generated branch we store its sample identifier, branch index, checkpoint location, remasking configuration, sampling seed, decoded answer, and correctness label. Branch correctness is evaluated using the same deterministic numeric evaluator described in Appendix~\ref{app:data_eval}. The strict branch label used by the verifier is the correctness of the branch under this common evaluator; auxiliary raw evaluation fields retained for diagnostics are not substituted for the formal training label.

The cached repair specification is also reused during branch hidden-state extraction. Rather than producing a new stochastic sample, the extractor reconstructs the same checkpoint, remasking configuration, branch seed, temperature, and sampling parameters. This allows the branch hidden trajectory to be paired with the exact branch whose answer and correctness label were stored in the repair pool.

\subsection{Branch-Trajectory Verifier}
\label{app:branch_verifier}

The branch verifier is a separate model from the first-stage wrong-lock-in detector. It uses the same underlying hidden-trajectory geometry described in Appendix~\ref{app:hidden_trajectory}, but is trained jointly on original trajectories and repair branches.

\paragraph{Branch trajectory representation.}
For each repair branch $b_i$, we select four relative positions
\[
\{0.25,0.50,0.75,1.00\}
\]
within the branch continuation and extract layers $5$, $10$, $15$, and $19$. The answer and change tracks, eight-token budgets, and 128-dimensional fixed random-sign projection are identical to Appendix~\ref{app:hidden_trajectory}. Consequently,
\[
\mathbf{H}^{\mathrm{br}}_i
\in
\mathbb{R}^{32768}.
\]
Branch replay is validated against the cached repair specification before the resulting hidden representation is retained.

\paragraph{Shared encoder and prediction heads.}
The second-stage model uses a shared trajectory encoder followed by two independent heads. The encoder has the same basic architecture as the detector in Appendix~\ref{app:gru_details}: each 128-dimensional trajectory element is projected to 256 dimensions and processed by a bidirectional GRU with hidden size 256 in each direction. The final forward and backward states are concatenated and layer normalized to obtain a 512-dimensional shared representation.

Two heads operate on this representation:
\[
f_{\phi}^{\mathrm{lock}}
\quad\text{and}\quad
f_{\phi}^{\mathrm{br}}.
\]
The first predicts correctness of the original lock-in trajectory and provides auxiliary supervision. The second outputs a scalar branch score
\[
v_i=
f_{\phi}^{\mathrm{br}}
\left(
\mathbf{H}^{\mathrm{br}}_i
\right),
\]
where higher scores indicate that a repair branch should be ranked above less promising alternatives. Each head uses the architecture
\[
512\rightarrow256\rightarrow1
\]
with GELU activation and dropout $0.30$.

\paragraph{Train-only lock-in pretraining.}
Before joint branch training, the shared encoder and lock-in head are pretrained using only lock-in examples assigned to the training partition. We use eight pretraining epochs in the formal configuration. This stage does not initialize from a model fitted on the complete matched benchmark; validation and test trajectories remain excluded.

\paragraph{Multi-task objective.}
Joint training combines three forms of supervision. First, the auxiliary lock-in loss is binary cross-entropy:
\[
\mathcal{L}_{\mathrm{lock}}
=
\operatorname{BCEWithLogits}
\left(
f_{\phi}^{\mathrm{lock}}(\mathbf{H}),
y_{\mathrm{lock}}
\right).
\]

Second, for a correct branch with score $v^{+}$ and an incorrect branch from the same training pool with score $v^{-}$, we use the pairwise ranking loss
\[
\mathcal{L}_{\mathrm{pair}}
=
\operatorname{softplus}
\left(
-\left(v^{+}-v^{-}-m\right)
\right),
\]
with margin
\[
m=0.
\]
This loss encourages correct branches to receive larger scores than incorrect alternatives without requiring the raw score itself to be calibrated as a probability.

Third, for a candidate set $\mathcal{B}_q$ belonging to one training example, let
\[
\mathcal{B}^{+}_q
=
\left\{
b_i\in\mathcal{B}_q:y_i=1
\right\}.
\]
For candidate sets containing both correct and incorrect branches, the listwise objective is
\[
\mathcal{L}_{\mathrm{list}}
=
\log
\sum_{b_i\in\mathcal{B}_q}
\exp(v_i)
-
\log
\sum_{b_i\in\mathcal{B}^{+}_q}
\exp(v_i).
\]
Minimizing this term increases the total score mass assigned to correct branches within each candidate set.

The complete training objective is
\[
\mathcal{L}
=
0.5\,\mathcal{L}_{\mathrm{lock}}
+
0.5\,\mathcal{L}_{\mathrm{pair}}
+
1.0\,\mathcal{L}_{\mathrm{list}}.
\]
Pairwise and listwise examples may additionally receive fixed source weights during training; these weights are determined before evaluation and do not depend on test performance.

\paragraph{Optimization.}
We use AdamW with learning rate $10^{-4}$ and weight decay $3\times10^{-4}$. Gradient norm is clipped at $5.0$. Joint training runs for at most 30 epochs with patience 6, following the validation-only model-selection protocol described in Appendix~\ref{app:training_validation}. The formal batches contain 256 lock-in examples, 128 correct--incorrect branch pairs, and 32 candidate sets for the listwise objective; evaluation uses batch size 512. The formal planner wrappers use seed 17.

\begin{table}[t]
\centering
\small
\caption{Multi-task branch-trajectory verifier settings.}
\label{tab:branch_verifier_config}
\begin{tabular}{ll}
\toprule
\textbf{Parameter} & \textbf{Setting} \\
\midrule
Trajectory dimension & 32,768 \\
Input projection & $128\rightarrow256$ \\
GRU hidden size & 256 per direction \\
Shared representation & 512 \\
Prediction heads & lock-in, branch \\
Dropout & 0.30 \\
Lock-in pretraining epochs & 8 \\
Maximum joint epochs & 30 \\
Early-stopping patience & 6 \\
Optimizer & AdamW \\
Learning rate & $1\times10^{-4}$ \\
Weight decay & $3\times10^{-4}$ \\
Gradient clipping & 5.0 \\
$\lambda_{\mathrm{lock}}$ & 0.5 \\
$\lambda_{\mathrm{pair}}$ & 0.5 \\
$\lambda_{\mathrm{list}}$ & 1.0 \\
Pairwise margin & 0 \\
Lock-in batch size & 256 \\
Pairwise batch size & 128 \\
Listwise batch size & 32 \\
Evaluation batch size & 512 \\
Training seed & 17 \\
\bottomrule
\end{tabular}
\end{table}

After training, the trajectory model is frozen. Its branch scores are computed once for the repair caches and subsequently treated as fixed features by the final selector.

\subsection{Final HGB Selector}
\label{app:hgb_selector}

The branch-trajectory score contains strong correctness information, but a repair candidate should also be interpreted relative to the original generation and to the other branches available for the same example. We therefore train a final histogram gradient-boosted classifier over branch-level features. This model is fitted only after the trajectory verifier has been frozen.

\paragraph{Feature construction.}
For branch $b_i$, the selector feature vector contains four broad groups of information.

\begin{enumerate}
    \item \textbf{Trajectory evidence.} The frozen branch-trajectory score $v_i$, together with within-candidate-set transformations such as its rank, candidate-set mean and standard deviation, gap from the highest branch score, and standardized within-set score.
    \item \textbf{First-stage risk context.} The original generation's frozen first-stage correctness score $f_{\theta}(\mathbf{H})$, wrong-lock-in risk $p_{\mathrm{risk}}=1-f_{\theta}(\mathbf{H})$, and its relative risk percentile.
    \item \textbf{Repair and stability metadata.} Branch index, nominal remasking budget, branch sampling metadata, number of recorded branch snapshots when available, and lightweight stability statistics inherited from the original trajectory.
    \item \textbf{Consensus and answer agreement.} The frequency of the branch answer within the candidate set, number of unique candidate answers, majority fraction, normalized answer entropy, whether the branch agrees with the majority answer, and whether it agrees with the baseline or locked answer. We additionally include within-set ranks, gaps, and standardized versions of these consensus quantities.
\end{enumerate}

Some branch caches retain auxiliary frozen branch-score fields produced during earlier branch construction. When present, these are treated as additional branch-level trajectory evidence; when absent, they remain missing. Features that are entirely missing or constant on the training partition are removed automatically before fitting. This allows the same selector implementation to operate across the two model families without introducing test-dependent feature choices.

\paragraph{Classifier.}
Let
\[
\mathbf{z}_i
\]
denote the resulting feature vector for candidate $b_i$. The selector produces
\[
s_i
=
g_{\psi}(\mathbf{z}_i),
\]
where $g_{\psi}$ is a histogram gradient-boosted classifier trained on branch correctness. We use log loss and class-balanced sample weights. Missing values are median imputed using the training data, followed by standardization before the gradient-boosted classifier.

The formal configuration uses 250 boosting iterations, learning rate $0.05$, at most 31 leaf nodes per tree, and $\ell_2$ regularization coefficient $0.10$. Internal early stopping uses a $10\%$ training-side validation fraction with patience 20. All randomization uses seed 17.

\begin{table}[t]
\centering
\small
\caption{Final histogram gradient-boosted branch selector settings.}
\label{tab:hgb_config}
\begin{tabular}{ll}
\toprule
\textbf{Parameter} & \textbf{Setting} \\
\midrule
Classifier & HistGradientBoostingClassifier \\
Loss & log loss \\
Maximum iterations & 250 \\
Learning rate & 0.05 \\
Maximum leaf nodes & 31 \\
$\ell_2$ regularization & 0.10 \\
Missing-value preprocessing & median imputation \\
Feature scaling & standardization \\
Class weighting & balanced \\
Internal validation fraction & 0.10 \\
No-improvement patience & 20 \\
Seed & 17 \\
\bottomrule
\end{tabular}
\end{table}

For a fixed repair pool, the selected branch is
\[
b^{\star}
=
\arg\max_{b_i\in\mathcal{B}} s_i.
\]
The held-out branch-selection comparisons in Section~\ref{sec:exp_branch_selection} keep $\mathcal{B}$ fixed and vary only this selection mechanism, so differences among random selection, model confidence, Surface-18, external PRM, single hidden snapshot representations, and \textsc{LOCKR} cannot be attributed to different candidate pools.

\paragraph{End-to-end selective repair.}
Branch selection is separated from the first-stage intervention decision. The first-stage detector assigns each original generation a wrong-lock-in risk, and the repair operating point is selected using validation data. Generations below the intervention threshold retain the baseline decoding path and incur no branch-generation cost. For a generation selected for repair, the fixed action bank is executed, branch trajectories are scored, and the HGB selector ranks the resulting candidates. The final test policy and its operating point are frozen before test evaluation.

This separation is important for interpreting the ablations in the main paper. The first-stage trajectory model primarily determines where additional computation is allocated, the fixed repair bank determines which counterfactual alternatives become available, and the second-stage trajectory verifier together with the HGB selector determines which available alternative is retained. Oracle branch selection removes only the final within-pool selection error by choosing a correct branch whenever one exists. The end-to-end repair oracle additionally preserves a correct baseline output and uses a correct repair for a baseline-wrong example whenever one is available. Neither oracle changes the underlying repair pool. Consequently, residual end-to-end error can be separated into whether a successful repair is available and whether the learned selector identifies it.

\section{Additional Detection Results}
\label{app:detection_results}

This appendix provides additional analyses of the first-stage SBW detector studied in Section~\ref{sec:exp_detection}. The main text reports a compact AUROC comparison for the principal surface and hidden-state representations. Table~\ref{tab:app_wrong_lockin_detection_full} provides the complete detector comparison, including scalar surface cues and matched-pair ranking accuracy. We then report additional thresholded and ranking metrics for the trajectory detector, clarify the validation-only construction of the single hidden snapshot baseline, and examine how wrong-lock-in concentration changes as the intervention coverage is varied.

\begin{table*}[t]
\centering
\scriptsize
\setlength{\tabcolsep}{3.4pt}
\caption{Full wrong-lock-in detection results under surface-matched evaluation. Each entry reports AUROC / pairwise accuracy (\%). Surface-18 LR and HGB use the same 18 observable decoding statistics used for matching, and the single hidden snapshot is selected using validation data only.}
\label{tab:app_wrong_lockin_detection_full}
\begin{tabular}{lccccc}
\toprule
& \multicolumn{3}{c}{\textbf{DiffusionGemma}} & \multicolumn{2}{c}{\textbf{LLaDA-2}} \\
\cmidrule(lr){2-4}\cmidrule(lr){5-6}
\textbf{Detector} & \textbf{MetaMathQA} & \textbf{Orca-Math} & \textbf{NuminaMath V2} & \textbf{MetaMathQA} & \textbf{Orca-Math} \\
\midrule
Confidence & 49.97 / 49.12 & 49.99 / 50.82 & 49.96 / 49.99 & 50.34 / 53.32 & 50.23 / 53.30 \\
Entropy & 50.39 / 56.71 & 50.60 / 57.61 & 50.46 / 54.02 & 51.09 / 56.05 & 50.90 / 53.70 \\
Top-2 margin & 50.10 / 50.88 & 50.08 / 52.47 & 49.88 / 48.73 & 50.31 / 51.67 & 50.23 / 51.56 \\
Stability / flip & 50.00 / 50.00 & 50.00 / 50.00 & 50.00 / 50.00 & 50.50 / 51.36 & 50.77 / 52.34 \\
Surface-18 LR & 62.91 / 67.03 & 64.90 / 69.06 & 60.36 / 66.15 & 53.61 / 56.85 & 54.39 / 57.65 \\
Surface-18 HGB & 60.76 / 61.25 & 64.25 / 66.99 & 62.38 / 63.85 & 56.31 / 57.60 & 56.09 / 57.14 \\
Single hidden snapshot & 81.32 / 81.47 & 77.46 / 77.39 & 75.26 / 76.48 & 82.59 / 82.66 & 75.04 / 76.47 \\
\textbf{\textsc{LOCKR} hidden trajectory} & \textbf{83.04 / 82.88} & \textbf{80.49 / 82.14} & \textbf{77.29 / 78.24} & \textbf{83.92 / 84.08} & \textbf{78.65 / 78.84} \\
\bottomrule
\end{tabular}
\end{table*}

\subsection{Full Detection Metrics}
\label{app:full_detection_metrics}

The first-stage trajectory model is trained to estimate early-lock-in correctness,
\[
f_{\theta}(\mathbf{H})
=
P_{\theta}
\left(
\text{early-correct lock-in}\mid\mathbf{H}
\right),
\]
and the planner uses the inverted score
\[
p_{\mathrm{risk}}
=
1-f_{\theta}(\mathbf{H})
\]
as wrong-lock-in risk. For clarity, all metrics in this subsection are reported using the wrong-lock-in convention: early-wrong examples are treated as the positive class and larger $p_{\mathrm{risk}}$ indicates greater predicted risk.

Table~\ref{tab:app_full_detection_metrics} reports AUROC, AUPRC, classification accuracy, F1, matched-pair ranking accuracy, and precision within the highest-risk $10\%$ of test examples. AUROC, AUPRC, and pairwise accuracy are threshold independent. Accuracy and F1 use the natural score midpoint $p_{\mathrm{risk}}=0.5$ only as a descriptive diagnostic; the end-to-end repair policy does not select its intervention threshold from the test set.

\begin{table*}[t]
\centering
\scriptsize
\setlength{\tabcolsep}{4.2pt}
\caption{Additional first-stage trajectory-detector metrics on the surface-matched test sets. All values are percentages. AUPRC, accuracy, and F1 treat SBW lock-in as the positive class. Pairwise accuracy measures whether the wrong member of a matched pair receives higher risk. Top-10\% precision is the fraction of wrong examples among the $10\%$ highest-risk test instances.}
\label{tab:app_full_detection_metrics}
\begin{tabular}{llrrrrrr}
\toprule
\textbf{Model} & \textbf{Dataset} & \textbf{AUROC} & \textbf{AUPRC} & \textbf{Acc.} & \textbf{F1} & \textbf{Pairwise} & \textbf{Top-10\% wrong prec.} \\
\midrule
DiffusionGemma & MetaMathQA & 83.04 & 84.09 & 74.28 & 74.41 & 82.88 & 98.50 \\
DiffusionGemma & Orca-Math & 80.49 & 79.98 & 72.86 & 72.46 & 82.14 & 91.91 \\
DiffusionGemma & NuminaMath V2 & 77.29 & 75.43 & 69.87 & 71.10 & 78.24 & 85.88 \\
LLaDA-2 & MetaMathQA & 83.92 & 83.74 & 75.57 & 76.60 & 84.08 & 94.12 \\
LLaDA-2 & Orca-Math & 78.65 & 76.82 & 71.52 & 72.27 & 78.84 & 86.62 \\
\bottomrule
\end{tabular}
\end{table*}

The additional metrics reinforce the ranking results in the main text. AUPRC remains between $75.43\%$ and $84.09\%$ across the five evaluated settings, showing that the detector's performance is not specific to AUROC. At a fixed midpoint threshold, accuracy ranges from $69.87\%$ to $75.57\%$, despite the correct and wrong classes being balanced by construction.

More importantly for selective repair, the highest-risk portion of the score distribution is strongly enriched for SBW examples. On DiffusionGemma + MetaMathQA, $98.5\%$ of the highest-risk $10\%$ of test examples are wrong lock-ins. The corresponding precision remains $91.9\%$ on DiffusionGemma + Orca-Math and $94.1\%$ on LLaDA-2 + MetaMathQA. Even on the more difficult DiffusionGemma + NuminaMath V2 and LLaDA-2 + Orca-Math settings, the highest-risk decile contains $85.9\%$ and $86.6\%$ wrong lock-ins, respectively. This concentration is particularly relevant to \textsc{LOCKR}, because the first-stage detector is used to allocate additional computation rather than to make a final binary correctness decision.

\subsection{Single Hidden Snapshot Selection}
\label{app:snapshot_selection}

The single hidden snapshot baseline is designed to isolate the benefit of temporal coverage from the benefit of access to hidden representations themselves. It uses the same hidden-state extraction geometry as Appendix~\ref{app:hidden_trajectory}, but retains only one of the four candidate relative prefix positions,
\[
\mathcal{R}
=
\{0.25,0.50,0.75,1.00\}.
\]
A single snapshot therefore contains
\[
4\times2\times8\times128
=
8192
\]
hidden features corresponding to four layers, two token tracks, eight positions per track, and 128 projected dimensions.

We train a separate single hidden snapshot detector for each candidate position. Model fitting, normalization, and checkpoint selection use only the training and validation partitions. The snapshot used for final comparison is selected by validation AUROC, with validation pairwise accuracy used to break ties. Only after this choice is frozen do we evaluate the selected snapshot on the held-out test pairs. We therefore never choose the observation position according to test performance.

Figure~\ref{fig:detection_hierarchy} provides the main-text visual summary of this surface-to-hidden hierarchy. Table~\ref{tab:app_snapshot_gain} reports the improvement of the complete trajectory representation over the validation-selected single hidden snapshot baseline. The underlying AUROC and pairwise values are reported in Appendix Table~\ref{tab:app_wrong_lockin_detection_full}.

\begin{table}[t]
\centering
\small
\caption{Gain from modeling the full hidden trajectory rather than a validation-selected single hidden snapshot. Values are absolute percentage-point improvements on the frozen test sets.}
\label{tab:app_snapshot_gain}
\begin{tabular}{lrr}
\toprule
\textbf{Setting} & \textbf{$\Delta$ AUROC (pp)} & \textbf{$\Delta$ Pairwise (pp)} \\
\midrule
DiffusionGemma + MetaMathQA & +1.72 & +1.41 \\
DiffusionGemma + Orca-Math & +3.03 & +4.75 \\
DiffusionGemma + NuminaMath V2 & +2.03 & +1.76 \\
LLaDA-2 + MetaMathQA & +1.33 & +1.42 \\
LLaDA-2 + Orca-Math & +3.61 & +2.37 \\
\bottomrule
\end{tabular}
\end{table}

The full trajectory improves both AUROC and pairwise accuracy in every evaluated setting. The improvement is modest but consistent on MetaMathQA, where a single hidden state already contains substantial correctness information, and becomes larger on Orca-Math. In particular, the pairwise improvement reaches $4.75$ percentage points for DiffusionGemma + Orca-Math, while the AUROC improvement reaches $3.61$ percentage points for LLaDA-2 + Orca-Math.

This comparison is intentionally favorable to the snapshot baseline: the observation position is not fixed arbitrarily, but selected separately for each model--dataset setting using validation performance. The resulting gain therefore reflects information contributed by modeling multiple stages of the hidden trajectory rather than a poor choice of a single observation point. Complete per-position validation logs are retained with the supplementary experimental artifacts.

\subsection{Operating-Point and Risk-Coverage Sensitivity}
\label{app:risk_coverage}

The first-stage detector produces a continuous wrong-lock-in risk score rather than a fixed intervention decision. Consequently, the amount of additional test-time computation can be adjusted by changing the operating point. Raw score thresholds are not directly comparable across independently trained model--dataset detectors, so we analyze sensitivity in terms of \emph{risk coverage}: the fraction of examples assigned to the highest-risk region.

For a coverage level $c\in(0,1]$, let
\[
\mathcal{T}_{c}
\]
denote the top $c$ fraction of test examples after sorting by $p_{\mathrm{risk}}$. We define
\[
P_{\mathrm{wrong}}(c)
=
\frac{
\sum_{i\in\mathcal{T}_{c}}
\mathbb{I}[y_i=\mathrm{wrong}]
}{
|\mathcal{T}_{c}|
}.
\]
Because the surface-matched test sets contain equal numbers of correct and wrong lock-ins, random selection would yield an expected wrong precision of $50\%$ at every coverage level.

Table~\ref{tab:app_risk_coverage} reports the observed precision as coverage increases from $5\%$ to $30\%$.

\begin{table*}[t]
\centering
\small
\caption{Wrong-lock-in precision among the highest-risk test examples as a function of risk coverage. All values are percentages. The balanced matched-set reference is $50\%$.}
\label{tab:app_risk_coverage}
\begin{tabular}{llrrrr}
\toprule
\textbf{Model} & \textbf{Dataset} & \textbf{Top 5\%} & \textbf{Top 10\%} & \textbf{Top 20\%} & \textbf{Top 30\%} \\
\midrule
DiffusionGemma & MetaMathQA & 98.5 & 98.5 & 92.0 & 85.7 \\
DiffusionGemma & Orca-Math & 95.0 & 91.9 & 86.8 & 81.5 \\
DiffusionGemma & NuminaMath V2 & 89.5 & 85.9 & 80.7 & 76.5 \\
LLaDA-2 & MetaMathQA & 96.0 & 94.1 & 90.6 & 85.9 \\
LLaDA-2 & Orca-Math & 90.0 & 86.6 & 82.6 & 78.2 \\
\bottomrule
\end{tabular}
\end{table*}

Risk enrichment decreases smoothly as the selected region expands, as expected, but remains substantial even at $30\%$ coverage. At this operating range, between $76.5\%$ and $85.9\%$ of selected examples are SBW, compared with the $50\%$ matched-set base rate. At lower coverage, the detector becomes highly selective: the top $5\%$ contains approximately $90$--$99\%$ wrong lock-ins across the five settings.

These results provide a direct interpretation of the detector as a compute-allocation signal. A conservative operating point can reserve repair for a small, high-precision risk set, whereas a lower threshold can increase wrong-lock-in recall at the cost of repairing more correct examples. The end-to-end \textsc{LOCKR} results do not choose among these operating points using test outcomes. Instead, the intervention operating point is selected using validation data as described in Appendix~\ref{app:training_validation}, frozen, and then applied to the test set and to the natural-distribution evaluation.

The sensitivity analysis therefore separates two questions. AUROC and pairwise accuracy measure whether the hidden trajectory orders correct and wrong lock-ins reliably across the full score range, while risk-coverage precision measures whether the extreme-risk tail is sufficiently concentrated to support selective intervention. \textsc{LOCKR} performs well under both views: the trajectory detector provides strong global ranking and, at the same time, identifies a substantially enriched subset of generations on which additional repair computation is most justified.

\section{Additional Repair Results}
\label{app:repair_results}

This appendix provides additional analyses of the repair and branch-selection experiments in Sections~\ref{sec:exp_branch_selection}--\ref{sec:exp_generalization}. We first report the complete DiffusionGemma end-to-end comparison, followed by the full fixed-pool selector results and branch-level discrimination. We then define the per-action diagnostics used to characterize the fixed repair bank, analyze repair-space coverage and end-to-end oracle headroom, and finally report additional control baselines omitted from the main text for compactness.

\subsection{Full End-to-End DiffusionGemma Comparison}
\label{app:full_e2e_diffusiongemma}

Appendix Table~\ref{tab:app_e2e_diffusiongemma_full} reports the complete end-to-end comparison corresponding to the compact main-text Table~\ref{tab:e2e_diffusiongemma}. All methods use the same diffusion-model denoising-cost accounting described in Appendix~\ref{app:cost_accounting}; auxiliary scoring-model compute such as external PRM forward passes is excluded from the reported relative cost.

\begin{table*}[t]
\centering
\small
\caption{Complete end-to-end test-time repair results on the surface-matched DiffusionGemma benchmark. Each entry reports final accuracy (\%) / relative diffusion-model denoising cost. Standard decoding is normalized to $1.0\times$ cost. The end-to-end repair oracle retains correct baseline outputs and returns a correct repair for baseline-wrong examples whenever the fixed repair pool contains one.}
\label{tab:app_e2e_diffusiongemma_full}
\begin{tabular}{lccc}
\toprule
\textbf{Method} & \textbf{MetaMathQA} & \textbf{Orca-Math} & \textbf{NuminaMath V2} \\
\midrule
Base decoding & 50.0 / 1.0 & 50.0 / 1.0 & 50.0 / 1.0 \\
Random-trigger repair & 53.2 / 2.4 & 51.6 / 2.4 & 48.7 / 2.2 \\
Full restart / Self-Consistency & 56.5 / 4.8 & 55.2 / 4.8 & 50.4 / 4.5 \\
Confidence-guided repair & 53.2 / 2.9 & 52.7 / 2.9 & 49.9 / 2.5 \\
Surface-18 planner & 60.8 / 2.7 & 61.1 / 2.8 & 53.6 / 2.2 \\
Always repair + Majority & 58.8 / 5.0 & 58.2 / 5.0 & 51.9 / 5.0 \\
ReMDM & 61.3 / 2.7 & 59.2 / 2.7 & 52.4 / 2.4 \\
PRM-guided repair & 61.9 / 6.4 & 60.2 / 6.4 & 52.8 / 6.2 \\
\textbf{\textsc{LOCKR}} & \textbf{67.9 / 3.1} & \textbf{66.4 / 3.1} & \textbf{55.4 / 2.9} \\
End-to-end repair oracle & 72.8 / -- & 72.9 / -- & 62.3 / -- \\
\bottomrule
\end{tabular}
\end{table*}

\begin{table*}[t]
\centering
\small
\setlength{\tabcolsep}{3pt}
\caption{Complete fixed-pool repair-branch selection accuracy (\%). All selectors operate on the same candidate branches. Oracle branch selection chooses a correct repair whenever at least one correct branch is available and therefore represents a branch-pool upper bound.}
\label{tab:app_branch_selection_full}
\begin{tabular*}{\textwidth}{@{\extracolsep{\fill}}lccccc@{}}
\toprule
& \multicolumn{3}{c}{\textbf{DiffusionGemma}} & \multicolumn{2}{c}{\textbf{LLaDA-2}} \\
\cmidrule(lr){2-4}\cmidrule(lr){5-6}
\textbf{Branch selector}
& \textbf{MetaMathQA}
& \textbf{Orca-Math}
& \shortstack{\textbf{NuminaMath}\\\textbf{V2}}
& \textbf{MetaMathQA}
& \textbf{Orca-Math} \\
\midrule
Random branch & 54.6 & 53.8 & 48.9 & 40.8 & 39.4 \\
Max confidence & 54.8 & 54.2 & 49.1 & 40.5 & 39.2 \\
Surface-18 verifier & 59.6 & 60.2 & 52.4 & 44.0 & 42.3 \\
External PRM & 62.3 & 60.1 & 52.5 & 48.8 & 45.9 \\
Single hidden snapshot & 64.5 & 62.9 & 53.6 & 51.7 & 47.7 \\
\textbf{\textsc{LOCKR}} & \textbf{67.8} & \textbf{66.2} & \textbf{55.2} & \textbf{56.0} & \textbf{50.8} \\
Oracle branch selection & 72.8 & 72.8 & 62.1 & 68.7 & 63.7 \\
\bottomrule
\end{tabular*}
\end{table*}

\subsection{Full Branch-Selection Metrics}
\label{app:full_branch_selection}

The fixed-pool experiment in Section~\ref{sec:exp_branch_selection} isolates branch-selection quality by presenting every selector with the same repair candidates. Appendix Table~\ref{tab:app_branch_selection_full} reports the complete selected-branch accuracy comparison, while the main text retains a compact subset of the strongest representative baselines. We additionally evaluate the continuous branch scores before the final within-pool argmax decision.

Let $s_i$ denote the final \textsc{LOCKR} selector score assigned to repair branch $b_i$, as defined in Appendix~\ref{app:hgb_selector}. Treating strict branch correctness as the positive label, branch-level AUROC measures whether the selector globally separates successful and unsuccessful repair trajectories. This metric differs from selected-branch accuracy: AUROC evaluates all individual branches pooled across examples, whereas selected-branch accuracy evaluates whether the highest-scoring branch within each candidate set is correct.

Table~\ref{tab:app_branch_full_metrics} reports both quantities together with the branch-pool oracle and the remaining selection gap.

\begin{table*}[t]
\centering
\small
\setlength{\tabcolsep}{3pt}
\caption{Additional fixed-pool branch-selection metrics. Selected-branch accuracy measures correctness of the highest-scoring repair candidate within each pool. Branch AUROC evaluates the continuous \textsc{LOCKR} selector scores over all repair candidates. Oracle branch selection chooses a correct repair whenever one exists in the fixed pool. The final column is the absolute gap between oracle branch-selection accuracy and \textsc{LOCKR} selected-branch accuracy. AUROC and accuracy values are percentages; the oracle gap is reported in percentage points.}
\label{tab:app_branch_full_metrics}
\begin{tabular*}{\textwidth}{@{\extracolsep{\fill}}llrrrr@{}}
\toprule
\textbf{Model}
& \textbf{Dataset}
& \shortstack{\textbf{\textsc{LOCKR} sel.}\\\textbf{acc.}}
& \shortstack{\textbf{Branch}\\\textbf{AUROC}}
& \shortstack{\textbf{Oracle branch}\\\textbf{acc.}}
& \shortstack{\textbf{Gap to}\\\textbf{oracle (pp)}} \\
\midrule
DiffusionGemma & MetaMathQA & 67.8 & 93.93 & 72.8 & 5.0 \\
DiffusionGemma & Orca-Math & 66.2 & 92.20 & 72.8 & 6.6 \\
DiffusionGemma & NuminaMath V2 & 55.2 & 86.34 & 62.1 & 6.9 \\
LLaDA-2 & MetaMathQA & 56.0 & 88.34 & 68.7 & 12.7 \\
LLaDA-2 & Orca-Math & 50.8 & 85.93 & 63.7 & 12.9 \\
\bottomrule
\end{tabular*}
\end{table*}

The selector achieves strong branch-level discrimination across all five settings, with AUROC ranging from $85.93\%$ to $93.93\%$. The remaining within-pool selection gap is substantially smaller for DiffusionGemma than for LLaDA-2. On DiffusionGemma, \textsc{LOCKR} remains within $5.0$--$6.9$ percentage points of oracle branch selection, whereas the corresponding gaps on LLaDA-2 are approximately $12.7$--$12.9$ percentage points. This indicates that the difficulty of repair differs along two dimensions: whether a correct branch is generated at all and, conditional on its availability, how reliably the selector can distinguish it from competing branches.

For DiffusionGemma + MetaMathQA, the underlying branch-level classifier reaches $93.93\%$ AUROC and the complete selector chooses a correct repair for $67.8\%$ of candidate sets, compared with a $72.8\%$ oracle branch-selection accuracy. This leaves a within-pool selection gap of only $5.0$ percentage points despite substantial variation in the correctness of individual repair branches. The corresponding gaps grow on harder model--dataset combinations, motivating the repair-space and verifier analyses below.

\subsection{Per-Action Repair Diagnostics}
\label{app:per_action_repair}

The formal repair bank contains eight paired checkpoint--budget actions,
\[
\mathcal{A}_i=(\rho_i,K_i),
\qquad
i\in\{1,\ldots,8\},
\]
with the exact configurations given in Table~\ref{tab:repair_action_bank}. Because checkpoint fraction and remasking budget are intentionally paired rather than varied independently, each action should be interpreted as one complete intervention configuration rather than as an isolated estimate of the effect of checkpoint location or remasking size.

We characterize every action using four complementary statistics. Let $y_q^{\mathrm{base}}\in\{0,1\}$ denote correctness of the original generation for example $q$, and let $y_{q,i}^{\mathrm{br}}\in\{0,1\}$ denote correctness of the branch generated by action $\mathcal{A}_i$.

\paragraph{Branch correctness.}
The unconditional correctness rate of action $i$ is
\[
\mathrm{Acc}_i
=
\frac{1}{N}
\sum_{q=1}^{N}
y_{q,i}^{\mathrm{br}}.
\]
This statistic measures the average quality of the branch generated by one repair configuration, but does not distinguish rescuing an originally wrong answer from preserving an originally correct one.

\paragraph{Wrong-lock-in rescue.}
For originally wrong examples, we define
\[
R_i^{\mathrm{wrong}}
=
P
\left(
y_{q,i}^{\mathrm{br}}=1
\mid
y_q^{\mathrm{base}}=0
\right).
\]
This is the most direct measure of whether an individual action can correct a SBW generation.

\paragraph{Correct-lock-in retention.}
For originally correct examples, we define
\[
R_i^{\mathrm{correct}}
=
P
\left(
y_{q,i}^{\mathrm{br}}=1
\mid
y_q^{\mathrm{base}}=1
\right).
\]
A repair action with high wrong-answer rescue but poor correct-answer retention may still be undesirable when applied without a selective trigger.

\paragraph{Unique repair-space contribution.}
Finally, we measure whether an action contributes a correct branch that would otherwise be absent from the repair pool:
\[
U_i
=
P
\left(
y_{q,i}^{\mathrm{br}}=1
\;\wedge\;
\sum_{j\neq i}y_{q,j}^{\mathrm{br}}=0
\right).
\]
This statistic distinguishes actions that merely duplicate alternatives already produced by other configurations from actions that expand the effective repair-space coverage of the bank.

The complete per-action tables, including $\mathrm{Acc}_i$, $R_i^{\mathrm{wrong}}$, $R_i^{\mathrm{correct}}$, $U_i$, answer diversity, and realized remasking metadata, are included in the anonymized supplementary artifacts for every formal model--dataset setting. We use the same fixed action ordering and correctness evaluator across settings. These diagnostics are descriptive: the eight actions themselves are fixed before training and are not selected or pruned using test-set performance.

This analysis also clarifies why the repair bank should be evaluated as a set rather than by choosing the single action with the highest marginal accuracy. The purpose of the bank is to expose complementary counterfactual continuations. An action can therefore be useful even when its standalone accuracy is not maximal, provided that it repairs examples for which other actions fail. The coverage analysis in the next subsection measures the combined reach of these alternatives.

\subsection{Repair Coverage and End-to-End Oracle Analysis}
\label{app:repair_coverage}

The following analyses separate failures caused by the repair space from failures caused by branch selection. We use oracle branch selection for the fixed-pool selection upper bound and the end-to-end repair oracle for the complete-policy upper bound. Let
\[
C_q
=
\mathbb{I}
\left[
\sum_{i=1}^{B}
y_{q,i}^{\mathrm{br}}>0
\right]
\]
indicate whether the fixed repair bank contains at least one correct branch for example $q$. We refer to
\[
P(C_q=1)
\]
as \emph{repair-space coverage}. Once a correct candidate exists, the remaining question is whether the learned selector ranks one of the successful branches first.

On the balanced surface-matched benchmark, the oracle branch-selection accuracies are $72.8\%$, $72.8\%$, and $62.1\%$ for DiffusionGemma on MetaMathQA, Orca-Math, and NuminaMath V2, respectively, and $68.7\%$ and $63.7\%$ for LLaDA-2 on MetaMathQA and Orca-Math. These values show that the same eight-action bank provides substantially different amounts of usable headroom across model--dataset combinations.

The LLaDA-2 results make this distinction particularly clear. Table~\ref{tab:app_llada_repair_coverage} reports the aggregate branch quality and the fraction of originally wrong examples for which at least one successful repair is exposed.

\begin{table}[t]
\centering
\small
\caption{Repair-space diagnostics for LLaDA-2. All reported performance values are percentages. Strict branch accuracy averages correctness over individual repair branches. Wrong-case repair coverage is the fraction of originally wrong lock-ins for which at least one of the eight repair actions produces a correct answer. The final column reports the end-to-end repair oracle, which retains correct baseline outputs and uses a correct repair whenever one is available for a baseline-wrong example.}
\label{tab:app_llada_repair_coverage}
\begin{tabular}{lrrrr}
\toprule
\textbf{Dataset} & \textbf{Strict branch acc.} & \textbf{Majority} & \textbf{Wrong-case cov.} & \textbf{E2E oracle} \\
\midrule
MetaMathQA & 40.51 & 41.71 & 43.33 & 71.66 \\
Orca-Math & 39.53 & 41.17 & 31.69 & 65.85 \\
\bottomrule
\end{tabular}
\end{table}

The average correctness of an individual repair branch is similar on the two LLaDA-2 datasets: $40.51\%$ on MetaMathQA and $39.53\%$ on Orca-Math. Majority voting is similarly close, at $41.71\%$ and $41.17\%$. The more consequential difference is repair coverage. Among originally wrong lock-ins, at least one action reaches the correct answer for $43.33\%$ of MetaMathQA cases but only $31.69\%$ of Orca-Math cases. The corresponding end-to-end repair oracle therefore falls from $71.66\%$ to $65.85\%$.

This result explains why strong detection and branch discrimination do not necessarily translate into equally large end-to-end improvements. The first-stage detector remains strong on LLaDA-2 + Orca-Math, and the branch selector still achieves substantial branch-level AUROC, but no selector can recover an example when every branch in its candidate pool is incorrect. Repair-space coverage therefore constitutes a hard ceiling for the current fixed action bank.

We can express the remaining error conceptually as two terms. First,
\[
1-P(C_q=1)
\]
captures failures of candidate generation: no successful counterfactual is exposed. Second, among examples satisfying $C_q=1$, the learned selector may fail to place a correct branch first. The latter produces the \emph{selection gap} quantified in Table~\ref{tab:app_branch_full_metrics}. This decomposition motivates two distinct future directions: richer or adaptive intervention policies can increase repair-space coverage, while stronger trajectory-aware verification can reduce the conditional selection error.

\subsection{Additional Baselines}
\label{app:additional_repair_baselines}

The main text emphasizes representative strong baselines in order to keep the end-to-end tables compact. We report additional weak and diagnostic controls here.

\paragraph{Majority voting on the fixed repair pool.}
Majority voting provides a selector that uses only answer agreement and no learned correctness signal. Its selected-branch accuracies are shown in Table~\ref{tab:app_majority_fixed_pool}.

\begin{table}[t]
\centering
\small
\caption{Majority-vote accuracy (\%) on the fixed repair pools. These results are omitted from the main fixed-pool selector table to avoid duplicating the always-repair majority baseline used in the end-to-end comparison.}
\label{tab:app_majority_fixed_pool}
\begin{tabular}{lrr}
\toprule
\textbf{Model} & \textbf{Dataset} & \textbf{Majority acc.} \\
\midrule
DiffusionGemma & MetaMathQA & 58.8 \\
DiffusionGemma & Orca-Math & 58.2 \\
DiffusionGemma & NuminaMath V2 & 51.9 \\
LLaDA-2 & MetaMathQA & 42.1 \\
LLaDA-2 & Orca-Math & 41.6 \\
\bottomrule
\end{tabular}
\end{table}

Majority voting provides a useful gain for DiffusionGemma on MetaMathQA and Orca-Math but is substantially weaker than trajectory-aware selection. On LLaDA-2, majority voting remains below the $50\%$ balanced-set baseline, illustrating that generating multiple repair branches does not by itself guarantee improvement.

\paragraph{Additional LLaDA-2 end-to-end controls.}
Table~\ref{tab:app_llada_weak_controls} reports three diagnostic controls omitted from the compact cross-model table in the main text. Each entry gives final accuracy and relative cost.

\begin{table}[t]
\centering
\small
\caption{Additional end-to-end controls on LLaDA-2. Each entry reports final accuracy (\%) / relative inference cost.}
\label{tab:app_llada_weak_controls}
\begin{tabular}{lcc}
\toprule
\textbf{Method} & \textbf{MetaMathQA} & \textbf{Orca-Math} \\
\midrule
Random-trigger repair & 40.7 / 2.3 & 38.8 / 2.1 \\
Always repair + Majority & 42.1 / 4.9 & 41.6 / 4.9 \\
Confidence-guided repair & 41.2 / 2.3 & 39.4 / 2.1 \\
\bottomrule
\end{tabular}
\end{table}

All three controls fail to improve over the $50\%$ balanced-set baseline on LLaDA-2. Random triggering demonstrates that allocating additional repair computation without a meaningful risk signal can actively harm performance. Confidence-guided repair behaves similarly, consistent with the weak surface-level discrimination observed in Section~\ref{sec:exp_detection}. Always repairing and then using majority voting incurs the largest cost among these controls while remaining substantially below the trajectory-aware planner.

Taken together, these additional results support the separation underlying \textsc{LOCKR}. Generating alternatives, identifying where repair is warranted, and selecting among the resulting candidates are distinct problems. Naive branch generation or answer aggregation alone is insufficient; the strongest performance is obtained when internal trajectories are used both to allocate repair computation and to evaluate the counterfactual trajectories that computation produces.

\section{Additional Ablations and Analysis}
\label{app:ablations}

This appendix extends the component and repair-space analyses in Section~\ref{sec:exp_ablation}. We first report additional ablations across model--dataset settings, then examine how selective intervention changes the accuracy--compute trade-off, and finally analyze the sources of residual error in the complete \textsc{LOCKR} pipeline. In contrast to Appendix~\ref{app:repair_results}, which focuses on the properties of the generated repair pool and branch selection, the analyses here focus on the behavior of the complete planning policy.

\subsection{Full Per-Dataset Ablations}
\label{app:full_ablations}

\paragraph{Component removal on DiffusionGemma.}
The main component study is conducted on DiffusionGemma + MetaMathQA, where the complete \textsc{LOCKR} policy reaches $67.9\%$ accuracy at $3.1\times$ relative cost. Table~\ref{tab:app_dg_component_ablation} reproduces the complete component-removal suite together with the full policy for reference.

\begin{table}[t]
\centering
\small
\caption{Complete component ablation on DiffusionGemma + MetaMathQA. $\Delta$Acc. and $\Delta$Cost are measured relative to full \textsc{LOCKR}; $\Delta$Acc. is reported in percentage points.}
\label{tab:app_dg_component_ablation}
\begin{tabular}{lrrrr}
\toprule
\textbf{Variant} & \textbf{Acc. (\%)} & \textbf{Cost} & \textbf{$\Delta$Acc. (pp)} & \textbf{$\Delta$Cost} \\
\midrule
Full \textsc{LOCKR} & 67.9 & 3.1 & -- & -- \\
Single hidden snapshot & 64.7 & 3.0 & -3.2 & -0.1 \\
w/o branch trajectory & 63.9 & 3.1 & -4.0 & 0.0 \\
w/o trajectory GRU & 63.7 & 2.9 & -4.2 & -0.2 \\
w/o targeted remasking & 62.7 & 2.9 & -5.2 & -0.2 \\
w/o adaptive trigger & 67.8 & 5.0 & -0.1 & +1.9 \\
Random trigger & 59.1 & 3.1 & -8.8 & 0.0 \\
\bottomrule
\end{tabular}
\end{table}

Several distinct roles emerge. Replacing the temporal hidden representation with a validation-selected single snapshot reduces final accuracy by $3.2$ percentage points, consistent with the trajectory advantage observed independently in detection and fixed-pool branch selection. Removing the branch-trajectory signal or the temporal GRU representation produces larger losses of $4.0$ and $4.2$ percentage points. Targeted remasking is also important: replacing the structured intervention with the corresponding non-targeted variant reduces accuracy by $5.2$ percentage points even though the overall inference cost changes only slightly.

The trigger ablations reveal a different role. Repairing every example recovers essentially the same accuracy as full \textsc{LOCKR}, but increases cost from $3.1\times$ to $5.0\times$. Conversely, replacing the learned risk signal with a random trigger at the same $3.1\times$ cost reduces accuracy by $8.8$ percentage points. Thus, the first-stage detector primarily controls \emph{where} computation is allocated, while targeted repair and trajectory-aware verification determine whether that additional computation is useful.

\paragraph{Cross-dataset verifier ablations on LLaDA-2.}
We additionally examine the second-stage trajectory components on LLaDA-2. These experiments use the formal V2 planner configuration and compare the complete trajectory-aware system with progressively weaker branch-verification variants. The fixed-pool verifier column measures selected-branch accuracy before the end-to-end trigger is applied, while the planner column reports final accuracy under the validation-selected repair policy. Because the LLaDA-2 formal planner uses a different ablation implementation from the DiffusionGemma study above, the variant names below follow the corresponding V2 components and are not intended as exact one-to-one counterparts of the DiffusionGemma removals.

\begin{table*}[t]
\centering
\small
\caption{Trajectory-verifier ablations on LLaDA-2. Branch AUROC evaluates branch-level discrimination, Verifier reports fixed-pool selected-branch accuracy, and Planner reports end-to-end accuracy on the balanced matched benchmark. All values are percentages.}
\label{tab:app_llada_ablation}
\begin{tabular}{llrrr}
\toprule
\textbf{Dataset} & \textbf{Variant} & \textbf{Branch AUROC} & \textbf{Verifier} & \textbf{Planner} \\
\midrule
MetaMathQA & Full \textsc{LOCKR} & 88.34 & 55.96 & 57.15 \\
& w/o branch GRU & 87.74 & 55.06 & 56.32 \\
& w/o GRU trajectory modeling & 81.48 & 50.38 & 51.73 \\
& Consensus only & 70.30 & 42.40 & 50.82 \\
\midrule
Orca-Math & Full \textsc{LOCKR} & 85.93 & 50.77 & 51.93 \\
& w/o branch GRU & 85.55 & 50.51 & 51.77 \\
& w/o GRU trajectory modeling & 83.03 & 49.48 & 50.57 \\
& Consensus only & 73.19 & 43.02 & 50.36 \\
\bottomrule
\end{tabular}
\end{table*}

The ordering is consistent across both LLaDA-2 datasets:
\[
\text{full trajectory}
>
\text{w/o branch GRU}
>
\text{w/o GRU}
>
\text{consensus only}.
\]
The magnitude of the end-to-end difference varies because the available repair headroom differs across datasets, but the branch-level results show that trajectory supervision contributes consistently to repair ranking. On MetaMathQA, removing GRU-based trajectory modeling decreases branch AUROC from $88.34\%$ to $81.48\%$ and reduces planner accuracy from $57.15\%$ to $51.73\%$. On Orca-Math, where the overall repair problem is more constrained, the same removal still reduces branch AUROC from $85.93\%$ to $83.03\%$.

These results complement the DiffusionGemma ablations: temporal hidden representations contribute both to identifying the original failure and to distinguishing the quality of counterfactual repair trajectories, while answer consensus alone provides substantially weaker supervision.

\subsection{Repair-Rate and Compute Sensitivity}
\label{app:compute_sensitivity}

\textsc{LOCKR} exposes a continuous intervention score, so inference cost can in principle be adjusted by changing the first-stage operating point. We do not select this operating point using test accuracy. Instead, the operating point is chosen on validation data and frozen before final evaluation, as described in Appendix~\ref{app:training_validation}. Figure~\ref{fig:accuracy_compute_tradeoff} provides the main-text view of the resulting matched-benchmark accuracy--compute trade-off. Here we analyze compute allocation from three complementary perspectives: concentration of errors in the high-risk tail, the realized repair rates on natural distributions, and the always-repair endpoint.

\paragraph{Risk concentration.}
Appendix~\ref{app:risk_coverage} shows that the wrong-lock-in precision of the selected risk set decreases smoothly as coverage increases. At $10\%$ coverage, between $85.9\%$ and $98.5\%$ of selected examples are SBW across the five evaluated settings. Even at $30\%$ coverage, wrong-lock-in precision remains between $76.5\%$ and $85.9\%$, substantially above the $50\%$ base rate of the balanced benchmark. This behavior indicates that the detector supports a continuum of operating points rather than relying on a narrowly tuned threshold.

\paragraph{Realized natural-distribution allocation.}
When the validation-selected policies are transferred to the original evaluation distributions, the resulting repair rates vary naturally with the model--dataset setting rather than being fixed in advance. Table~\ref{tab:app_natural_compute} summarizes the realized allocation and introduces an additional normalized efficiency statistic,
\[
\eta
=
\frac{
A_{\mathrm{\textsc{LOCKR}}}-A_{\mathrm{base}}
}{
C_{\mathrm{rel}}-1
},
\]
which measures percentage-point accuracy gain per additional base-generation unit of normalized inference work.

\begin{table*}[t]
\centering
\small
\caption{Realized compute allocation on the natural evaluation distributions. $\eta$ reports accuracy gain in percentage points per additional unit of normalized inference cost.}
\label{tab:app_natural_compute}
\begin{tabular}{llrrrr}
\toprule
\textbf{Model} & \textbf{Dataset} & \textbf{Repair rate} & \textbf{Rel. cost} & \textbf{$\Delta$ Acc. (pp)} & \textbf{$\eta$} \\
\midrule
DiffusionGemma & MetaMathQA & 22\% & 1.8 & +3.69 & 4.61 \\
DiffusionGemma & Orca-Math & 32\% & 2.2 & +4.22 & 3.52 \\
DiffusionGemma & NuminaMath V2 & 40\% & 2.6 & +4.20 & 2.63 \\
LLaDA-2 & MetaMathQA & 35\% & 2.4 & +5.37 & 3.84 \\
LLaDA-2 & Orca-Math & 41\% & 2.6 & +2.21 & 1.38 \\
\bottomrule
\end{tabular}
\end{table*}

The frozen policies repair only $22$--$41\%$ of natural examples, producing relative costs of $1.8$--$2.6\times$. The variation in repair rate reflects differences in the score distributions across model--dataset settings rather than test-time tuning to a target budget. In particular, the more difficult settings generally receive more intervention, while a relatively easy DiffusionGemma + MetaMathQA distribution requires repair on only $22\%$ of examples.

The efficiency statistic also shows that equal additional compute does not imply equal realizable benefit. DiffusionGemma + MetaMathQA gains $4.61$ percentage points of accuracy per additional normalized compute unit, whereas LLaDA-2 + Orca-Math gains only $1.38$ percentage points. As examined in Appendix~\ref{app:failure_analysis}, this difference is primarily associated with the amount of recoverable repair headroom rather than with an inability to identify risky examples.

\paragraph{Selective versus uniform intervention.}
The DiffusionGemma + MetaMathQA ablation provides a direct endpoint comparison. Full \textsc{LOCKR} reaches $67.9\%$ accuracy at $3.1\times$ cost, whereas repairing every example reaches $67.8\%$ at $5.0\times$. Thus, increasing intervention all the way to the always-repair regime adds $1.9\times$ relative cost without improving accuracy. At the same $3.1\times$ cost as \textsc{LOCKR}, a random trigger reaches only $59.1\%$ accuracy.

\begin{table}[t]
\centering
\small
\caption{Compute-allocation endpoints on DiffusionGemma + MetaMathQA. The comparison separates the amount of repair computation from the quality of its allocation.}
\label{tab:app_compute_endpoints}
\begin{tabular}{lrr}
\toprule
\textbf{Policy} & \textbf{Accuracy (\%)} & \textbf{Rel. cost} \\
\midrule
Base decoding & 50.0 & 1.0 \\
Random trigger & 59.1 & 3.1 \\
\textsc{LOCKR} & 67.9 & 3.1 \\
Always repair & 67.8 & 5.0 \\
\bottomrule
\end{tabular}
\end{table}

Together, these analyses show that the benefit of the first-stage detector is not merely to increase the amount of test-time computation. It concentrates repair on examples with a substantially elevated probability of being wrong, allowing \textsc{LOCKR} to approach the accuracy of uniform repair while using considerably less model-generation work.

\subsection{Failure and Repair-Space Analysis}
\label{app:failure_analysis}

The complete \textsc{LOCKR} pipeline can fail at several distinct stages. Separating these failure modes helps explain why detector quality alone does not determine the final gain.

Let $y_q^{\mathrm{base}}$ and $y_q^{\mathrm{final}}$ denote baseline and final correctness. The net accuracy change can be written as
\[
A_{\mathrm{final}}-A_{\mathrm{base}}
=
R-D,
\]
where
\[
R
=
P
\left(
y_q^{\mathrm{base}}=0,
y_q^{\mathrm{final}}=1
\right)
\]
is the rescue probability and
\[
D
=
P
\left(
y_q^{\mathrm{base}}=1,
y_q^{\mathrm{final}}=0
\right)
\]
is the damage probability. A successful planner must therefore both expose and select useful repairs for originally wrong examples while avoiding unnecessary degradation of originally correct generations.

We distinguish three major sources of residual error.

\paragraph{Risk-allocation error.}
The first-stage detector may fail to trigger on a wrong generation or may allocate repair to a correct generation. The high AUROC and high-risk precision results in Appendix~\ref{app:detection_results} show that this component is relatively strong, but it is not perfect. The random-trigger ablation demonstrates that the identity of the repaired examples matters substantially even when total compute is held fixed.

\paragraph{Repair-space failure.}
A triggered example cannot be corrected when none of the eight actions produces a correct counterfactual branch. This is a hard limitation of the fixed action bank and cannot be resolved by a better selector. The LLaDA-2 results provide a particularly clear example. Individual branch accuracy is similar on MetaMathQA and Orca-Math, at $40.51\%$ and $39.53\%$, respectively, yet wrong-case repair coverage decreases from $43.33\%$ to $31.69\%$. The corresponding end-to-end repair oracle falls from $71.66\%$ to $65.85\%$. Thus, many additional Orca-Math errors arise because the repair space never reaches the correct answer, rather than because all generated branches become uniformly lower quality.

\paragraph{Selection error.}
When the repair pool contains at least one correct branch, the learned verifier can still fail to rank a successful candidate first. Appendix~\ref{app:full_branch_selection} shows that this selection gap is approximately $5.0$--$6.9$ percentage points for DiffusionGemma and $12.7$--$12.9$ percentage points for LLaDA-2. These gaps indicate that both candidate generation and verification remain meaningful targets for improvement.

The cross-setting results illustrate how these stages interact. DiffusionGemma + MetaMathQA combines an $83.04\%$ wrong-lock-in detector AUROC with a $93.93\%$ branch AUROC and substantial repair headroom, yielding a large realized gain. DiffusionGemma + NuminaMath V2 remains detectable, but branch discrimination and the headroom under the end-to-end repair oracle are both lower, reducing the achievable improvement. LLaDA-2 + Orca-Math provides the clearest limiting case: the detector still reaches $78.65\%$ AUROC and the final branch selector reaches $85.93\%$ AUROC, yet the final improvement is small because the fixed repair bank reaches a correct alternative for substantially fewer wrong-lock-in examples.

A post-hoc operating-point diagnostic further supports this interpretation. On LLaDA-2 + Orca-Math, the formal validation-selected planner reaches $51.93\%$ test accuracy, while an exhaustive test-side operating-point grid has an upper bound of only $52.14\%$. The difference of $0.21$ percentage points indicates that the modest gain in this setting cannot be explained primarily by an unfortunate validation-selected threshold. Instead, the dominant limitation lies downstream, particularly in the coverage of the available repair space.

These observations suggest a natural decomposition of future improvements. Better risk models can reduce allocation errors, stronger trajectory-aware selectors can close the conditional selection gap, and adaptive repair-space construction can increase the probability that at least one successful counterfactual is generated. The current results indicate that the third direction becomes increasingly important once failure detection is already reasonably strong.

\section{Qualitative Examples}
\label{app:qualitative_examples}

This appendix provides representative held-out examples illustrating the trajectory patterns and repair outcomes studied throughout the paper. All four examples are drawn from the DiffusionGemma + MetaMathQA test split after completion of model training and quantitative evaluation. They are selected only for qualitative illustration and do not affect training, validation, operating-point selection, or reported metrics.

We consider four cases: a correct early lock-in, a SBW early lock-in, a successful \textsc{LOCKR} repair in which only one repair action reaches the correct answer, and an unsuccessful intervention for which the entire fixed repair bank fails to expose a correct branch. The early-lock-in category in each example is determined using the checkpoint and preceding stability window defined in Section~\ref{sec:early_lockin}.

\subsection{Correct Early Lock-In}
\label{app:qual_correct_lockin}

We first show a benign case in which early answer stability corresponds to a correct solution. The problem is
\[
\text{Find }x\text{ such that }\sqrt{3x+7}=10.
\]
The gold answer is $31$, since $3x+7=100$. DiffusionGemma exposes this answer immediately and retains it throughout all nine recorded answer observations.

\begin{table}[t]
\centering
\small
\caption{Representative correct early-lock-in example from DiffusionGemma + MetaMathQA.}
\label{tab:qual_correct_lockin}
\begin{tabular}{p{0.30\columnwidth}p{0.64\columnwidth}}
\toprule
\textbf{Field} & \textbf{Value} \\
\midrule
Sample ID & \path{metamathqa_00008398_15ee9328b2fe} \\
Gold answer & $31$ \\
Locked answer & $31$ \\
Baseline final answer & $31$ \\
Lock-in window & observations $2$--$4$ \\
Answer trajectory & $31\rightarrow31\rightarrow\cdots\rightarrow31$ \; (9 observations) \\
Mean confidence & $0.9755$ \\
Mean entropy & $0.1170$ \\
Surface-18 LR wrong risk & $0.5724$ \\
Hidden-trajectory wrong risk & $\mathbf{0.0002}$ \\
\bottomrule
\end{tabular}
\end{table}

The example is highly stable at the surface level: the answer does not change, the mean confidence is $0.9755$, and the mean entropy is low. Stability itself therefore provides no reason to regard this generation as pathological. Interestingly, the Surface-18 logistic baseline assigns a wrong-lock-in risk of $0.5724$, close to an ambiguous decision, whereas the hidden-trajectory detector assigns only $2.06\times10^{-4}$ wrong risk. The trajectory detector therefore assigns this apparently stable generation very low wrong-lock-in risk, whereas the observable statistics alone are less decisive.

This case illustrates the role of the first-stage detector on benign convergence. \textsc{LOCKR} is not designed to treat early stability as a failure signal by itself; rather, it uses the hidden evolution underlying that stable output to distinguish correct convergence from the SBW behavior illustrated next.

\subsection{SBW Lock-In}
\label{app:qual_sbw}

We next consider the problem:

\begin{quote}
When the decimal point of a certain positive decimal number is moved four places to the right, the new number is four times the reciprocal of the original number. What is the original number?
\end{quote}

If the original number is $x$, the condition gives
\[
10^4x=\frac{4}{x},
\]
and hence
\[
x^2=4\times10^{-4},
\qquad
x=0.02.
\]

DiffusionGemma instead settles on $0.2$ and retains this incorrect value through the remainder of the recorded trajectory.

\begin{table}[t]
\centering
\small
\caption{Representative SBW early-lock-in example.}
\label{tab:qual_sbw}
\begin{tabular}{p{0.30\columnwidth}p{0.64\columnwidth}}
\toprule
\textbf{Field} & \textbf{Value} \\
\midrule
Sample ID & \path{metamathqa_00185450_6fdb476232a7} \\
Gold answer & $0.02$ \\
Locked answer & $0.2$ \\
Baseline final answer & $0.2$ \\
Lock-in window & observations $3$--$6$ \\
Answer trajectory & $0\rightarrow0.5\rightarrow0.2\rightarrow\cdots\rightarrow0.2$ \\
Mean confidence & $0.9691$ \\
Mean entropy & $0.0929$ \\
Surface-18 LR wrong risk & $0.5227$ \\
Hidden-trajectory wrong risk & $\mathbf{0.999996}$ \\
\bottomrule
\end{tabular}
\end{table}

The surface behavior is deceptively similar to the correct example above. Once $0.2$ appears, the answer remains unchanged, mean confidence is $0.9691$, and mean entropy is only $0.0929$. The Surface-18 logistic model again remains close to an ambiguous decision, assigning wrong risk $0.5227$. In contrast, the hidden-trajectory model assigns wrong risk $0.999996$.

The contrast between Tables~\ref{tab:qual_correct_lockin} and~\ref{tab:qual_sbw} captures the central ambiguity motivating the surface-matched benchmark. Both trajectories appear stable and confident at the decoded level, yet the hidden representations place them at opposite ends of the predicted correctness spectrum.

The SBW label is determined at the intermediate checkpoint. It does not require an incorrect answer to remain unchanged throughout every later denoising step, although this particular example happens to preserve $0.2$ through the final output.

\subsection{Successful Repair and Trajectory-Aware Branch Selection}
\label{app:qual_successful_repair}

\begin{figure*}[t]
\centering
\includegraphics[width=0.94\textwidth]{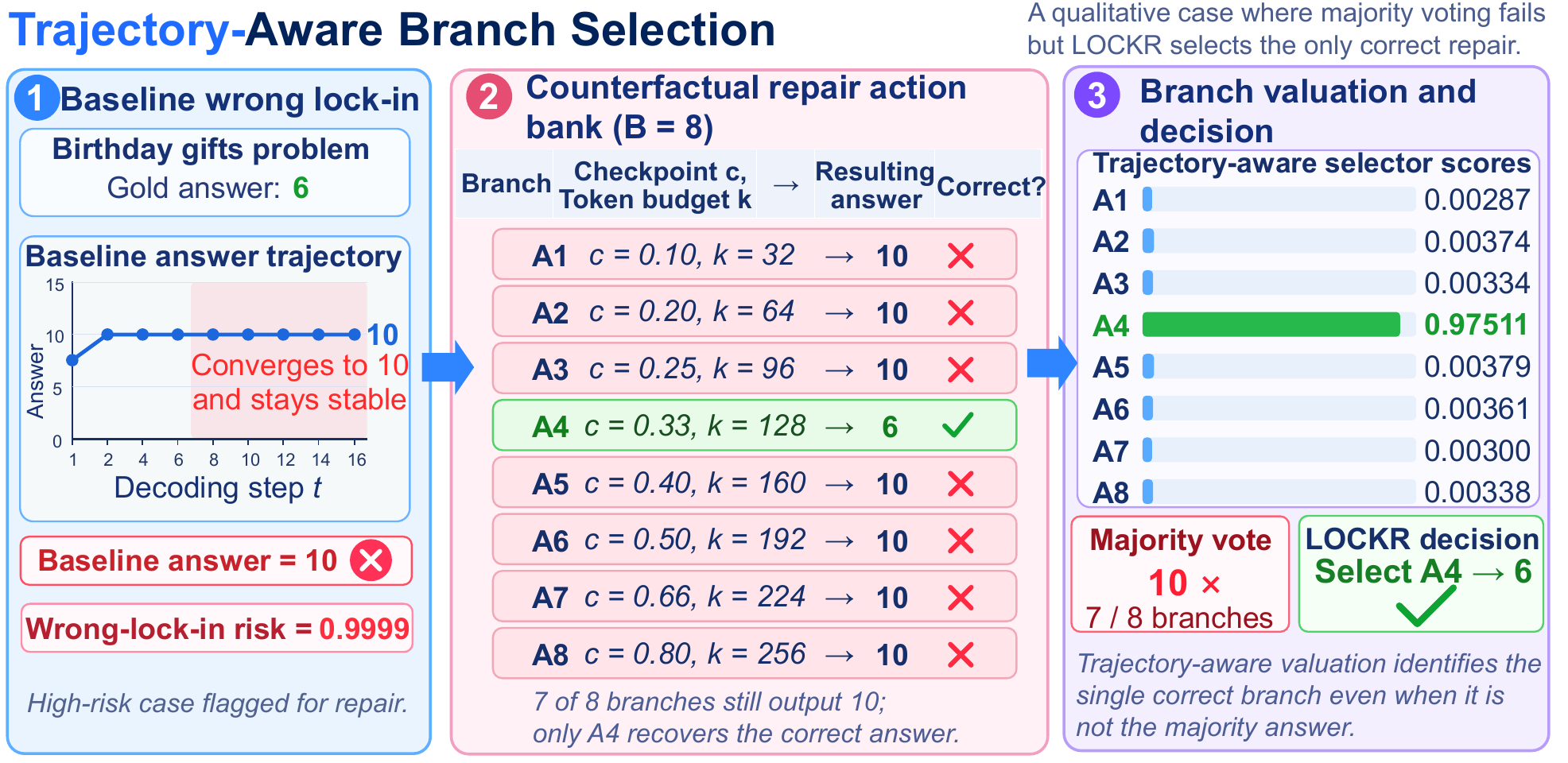}
\caption{\textbf{Trajectory-aware branch selection can recover a unique successful repair against an incorrect majority.} The example has gold answer $6$, while the baseline stabilizes on $10$ and triggers repair with wrong-lock-in risk $0.9999$. Seven of the eight fixed repair branches return $10$, whereas only Action~4 (checkpoint fraction $0.33$, remasking budget $128$) reaches $6$. Majority voting therefore fails, but the final \textsc{LOCKR} selector assigns Action~4 score $0.97511$ and retains the unique correct branch. Exact branch-level trajectory and selector scores are reported in Table~\ref{tab:qual_successful_repair_branches}.}
\label{fig:qual_branch_selection}
\end{figure*}

Figure~\ref{fig:qual_branch_selection} provides a visual summary of the following held-out DiffusionGemma + MetaMathQA example; the accompanying analysis and Table~\ref{tab:qual_successful_repair_branches} provide the exact problem, branch-level trajectory scores, and final selector scores.

The example illustrates the complete \textsc{LOCKR} pipeline:

\begin{quote}
Sixteen boys and 14 girls attended Simon's birthday party. Three-fourths of the boys and $6/7$ of the girls brought gifts. How many of those who attended did not bring gifts?
\end{quote}

Among the boys,
\[
16\times\frac{3}{4}=12
\]
bring gifts, and among the girls,
\[
14\times\frac{6}{7}=12
\]
bring gifts. Therefore
\[
30-(12+12)=6
\]
attendees do not bring gifts.

The baseline generation nevertheless transitions from an initial answer of $16$ to $10$ and then remains locked at $10$:
\[
16\rightarrow10\rightarrow10\rightarrow\cdots\rightarrow10.
\]
The lock-in window covers observations $3$--$6$, the final baseline answer is also $10$, and the hidden-trajectory detector assigns wrong risk $0.999922$. The Surface-18 logistic detector is substantially less decisive, assigning wrong risk $0.6015$.

Table~\ref{tab:qual_successful_repair_branches} shows all eight repair branches produced by the fixed action bank.

\begin{table*}[t]
\centering
\small
\setlength{\tabcolsep}{4.5pt}
\caption{Complete repair pool for a successful \textsc{LOCKR} intervention. The trajectory score is produced by the frozen branch-trajectory model, while the selector score is the final HGB score used for within-pool ranking. Answer frac. is the fraction of repair candidates assigned the same extracted-answer value. The only correct branch is highlighted in bold.}
\label{tab:qual_successful_repair_branches}
\begin{tabular}{crrrlrrc}
\toprule
\textbf{Action} & \textbf{Checkpoint} & \textbf{Budget} & \textbf{Answer frac.} & \textbf{Answer} & \textbf{Trajectory score} & \textbf{Selector score} & \textbf{Correct?} \\
\midrule
1 & 0.10 & 32  & 0.875 & $10$ & $0.000011$ & $0.00287$ & No \\
2 & 0.20 & 64  & 0.875 & $10$ & $0.000212$ & $0.00374$ & No \\
3 & 0.25 & 96  & 0.875 & $10$ & $0.000926$ & $0.00334$ & No \\
\textbf{4} & \textbf{0.33} & \textbf{128} & \textbf{0.125} & $\mathbf{6}$ & $\mathbf{0.999994}$ & $\mathbf{0.97511}$ & \textbf{Yes} \\
5 & 0.40 & 160 & 0.875 & $10$ & $0.000481$ & $0.00379$ & No \\
6 & 0.50 & 192 & 0.875 & $10$ & $0.000092$ & $0.00361$ & No \\
7 & 0.66 & 224 & 0.875 & $10$ & $0.000020$ & $0.00300$ & No \\
8 & 0.80 & 256 & 0.875 & $10$ & $0.000072$ & $0.00338$ & No \\
\bottomrule
\end{tabular}
\end{table*}

This example provides a particularly clear illustration of why repair generation and branch selection must be treated as separate problems. Seven of the eight repair actions reproduce the same incorrect answer $10$. The correct answer $6$ appears only under Action~4, corresponding to checkpoint fraction $0.33$ and nominal remasking budget $128$. Consequently, a simple majority vote would retain the incorrect answer:
\[
P_{\mathrm{pool}}(a=10)=\frac{7}{8},
\qquad
P_{\mathrm{pool}}(a=6)=\frac{1}{8}.
\]

The branch trajectory tells a very different story. The correct branch receives trajectory score $0.999994$, while all seven incorrect branches receive scores below $10^{-3}$. After combining this trajectory evidence with the other branch-level features, the final HGB selector assigns Action~4 a score of $0.97511$, whereas every incorrect alternative receives a score below $0.004$. The unique correct branch is therefore ranked first and selected:
\[
10
\quad\longrightarrow\quad
6.
\]

This case demonstrates the full mechanism intended by \textsc{LOCKR}: the prefix trajectory identifies a highly suspicious lock-in, targeted remasking exposes a rare successful counterfactual, and the repair trajectory allows the selector to recover that candidate even though it is strongly outvoted by the incorrect majority.

\subsection{Unsuccessful Repair: No Correct Branch in the Repair Pool}
\label{app:qual_unsuccessful_repair}

The final example illustrates the complementary failure mode in which detection succeeds but the fixed repair space never reaches the correct solution:

\begin{quote}
Dina has twice as many dolls as Ivy. $2/3$ of Ivy's dolls are collectors editions. If Ivy has 20 collectors edition dolls, how many dolls does Dina have?
\end{quote}

Since
\[
\frac{2}{3}I=20,
\]
Ivy has
\[
I=30
\]
dolls, and Dina therefore has
\[
2I=60.
\]

The baseline generation instead outputs $40$ at every recorded observation:
\[
40\rightarrow40\rightarrow\cdots\rightarrow40,
\]
with mean confidence $0.99935$ and mean entropy $0.00581$. Despite this extreme surface stability, the hidden-trajectory detector assigns wrong-lock-in risk $0.999994$. The Surface-18 logistic detector assigns only $0.5146$ wrong risk.

The complete repair pool is shown in Table~\ref{tab:qual_failed_repair_branches}.

\begin{table*}[t]
\centering
\small
\setlength{\tabcolsep}{4.5pt}
\caption{Complete repair pool for a repair-space failure. None of the eight fixed actions reaches the gold answer $60$. Answer frac. is the fraction of repair candidates assigned the same extracted-answer value, and a dash denotes a branch for which no valid numeric answer is extracted.}
\label{tab:qual_failed_repair_branches}
\begin{tabular}{crrrlrrc}
\toprule
\textbf{Action} & \textbf{Checkpoint} & \textbf{Budget} & \textbf{Answer frac.} & \textbf{Answer} & \textbf{Trajectory score} & \textbf{Selector score} & \textbf{Correct?} \\
\midrule
1 & 0.10 & 32  & 0.875 & $40$ & $0.002586$ & $0.00301$ & No \\
2 & 0.20 & 64  & 0.875 & $40$ & $0.000854$ & $0.00324$ & No \\
3 & 0.25 & 96  & 0.125 & --   & $0.000053$ & $0.00399$ & No \\
4 & 0.33 & 128 & 0.875 & $40$ & $0.005961$ & $0.00450$ & No \\
5 & 0.40 & 160 & 0.875 & $40$ & $0.001130$ & $0.00327$ & No \\
6 & 0.50 & 192 & 0.875 & $40$ & $0.007130$ & $0.00328$ & No \\
7 & 0.66 & 224 & 0.875 & $40$ & $0.010742$ & $0.00359$ & No \\
8 & 0.80 & 256 & 0.875 & $40$ & $0.003179$ & $0.00318$ & No \\
\bottomrule
\end{tabular}
\end{table*}

Seven branches reproduce the incorrect answer $40$, while the remaining branch does not yield a valid numeric answer. Thus,
\[
\sum_{i=1}^{8} y_i^{\mathrm{br}}=0,
\]
and the branch-pool oracle itself cannot recover this example. The final selector ranks Action~4 first, but this choice remains incorrect because every available candidate is incorrect.

This case separates \emph{failure detection} from \emph{repairability}. The original SBW trajectory is identified with extremely high risk, so the limitation does not originate in the first-stage trigger. Nor can it be solved by an improved selector, because no successful branch exists to select. The failure instead lies in the coverage of the fixed intervention space.

Taken together, the four examples illustrate the three distinct decisions underlying \textsc{LOCKR}. Hidden prefix trajectories distinguish benign stable convergence from SBW lock-in; structured interventions determine whether a successful counterfactual becomes available; and repair trajectories help identify that counterfactual when it exists. The successful example shows that trajectory-aware verification can recover a unique correct branch even against a $7$-to-$1$ incorrect majority, while the unsuccessful example makes explicit the remaining boundary of the method: no selector can recover a solution that the repair process never generates.

\end{document}